\documentclass[10pt,journal,twocolumn,twoside]{IEEEtran} 
\usepackage{cuted}            
\usepackage{color}            
\usepackage{cite}
\usepackage{bm}               
\usepackage{array}            
\usepackage{graphicx}         
\usepackage{algorithm,algorithmic}        
\usepackage{amsmath, amssymb, amsfonts, mathrsfs}   
\usepackage{amsthm}
\usepackage{booktabs}         
\usepackage{stfloats}         
\usepackage{fixltx2e}        
\usepackage{cases}            
\usepackage{mathtools}        
\usepackage{footmisc}         
\usepackage{epstopdf}
\usepackage[T1]{fontenc}      
\usepackage{titlesec}
\usepackage{makecell}         
\usepackage{subfig}           
\usepackage{caption}          
\usepackage{balance}
\usepackage{diagbox}
\usepackage{url}
\usepackage{pifont}
\usepackage{upgreek}
\usepackage{dsfont}
\newcommand\figref[1]{Fig.~\ref{#1}}

\newcommand{\tabref}[1]{Table~\ref{#1}}

\begin{document}

\title{Goal-Oriented Semantic Communication for Distributed ISAC-Enabled Vehicle Coordination}

\author{\IEEEauthorblockN{Wenjie Liu and Yansha Deng, \IEEEmembership{Senior Member, IEEE}}

\thanks{Wenjie Liu and Yansha Deng are with the Department of Engineering, King’s College London, Strand, London WC2R 2LS, U.K. (e-mail: wenjie.liu@kcl.ac.uk; yansha.deng@kcl.ac.uk) (Corresponding author: Yansha Deng).}
}
	
\maketitle

\begin{abstract}
Vehicle coordination at unsignalized intersections relies on accurate real-time vehicle state acquisition and reliable command-and-control (C\&C) signal delivery. However, existing studies typically treat sensing, communication, and control separately, which may lead to redundant transmissions, outdated state information, and unreliable vehicle coordination. In this paper, we investigate a new scenario of distributed integrated sensing and communication (ISAC)-enabled vehicle coordination at intersections, where multiple roadside units (RSUs) collaboratively transmit sensing signals for vehicle state acquisition and C\&C signals for vehicle movement control under the management of a central base station (BS). To improve signaling efficiency, we propose a unified goal-oriented semantic communication (GSC) framework, which transmits sensing and C\&C signals only when they are semantically important for improving intersection traffic throughput. Specifically, an extended Kalman filter (EKF) is adopted to predict vehicle states and fuse distributed sensing measurements. A masked hybrid proximal policy optimization (MHPPO) framework is then developed to jointly determine sensing transmission decisions, C\&C transmission decisions, and C\&C signal contents based on a value-of-information (VoI) reward. Furthermore, we propose an uncertainty-aware transmission design (UTD), including robust beamforming and VoI-based time-division power allocation, to improve sensing and communication reliability under vehicle state uncertainty and inter-RSU interference. Simulation results show that our proposed framework achieves 100\% collision-free vehicle coordination with significantly reduced signaling overhead compared with predictive ISAC baselines adapted from state-of-the-art related studies and several ablation baselines.
\end{abstract}

\section{Introduction}
\IEEEPARstart{V}{ehicle} coordination at unsignalized intersections has attracted widespread attention in intelligent transportation systems due to its potential to improve traffic efficiency \cite{UI}. Unlike traditional traffic signal control systems that typically rely on predetermined timing plans derived from historical statistics, unsignalized intersections generally employ an intersection manager to coordinate vehicle movements based on the real-time traffic conditions, thereby enabling more flexible and adaptive traffic management.

Recent studies on vehicle coordination at unsignalized intersections mainly focus on trajectory optimization \cite{traj}, priority scheduling \cite{priori}, and conflict-resolution mechanism design \cite{cr}. These methods improve intersection throughput, reduce travel delay, and enhance safety by optimizing vehicle movements and resolving potential conflicts. Despite their promising performance, existing studies generally rely on two fundamental premises: accurate real-time acquisition of vehicle states (e.g., position and velocity) and reliable low-latency command and control (C\&C) signal transmission for vehicle movement coordination. However, how to efficiently and jointly realize these two functions remains largely overlooked. In practice, implementing sensing and communication separately can cause spectrum inefficiency and high hardware costs. Moreover, insufficient coordination between sensing and communication may result in outdated vehicle state information and delayed C\&C delivery, thereby degrading the reliability and responsiveness of intersection management systems.

Integrated sensing and communication (ISAC) \cite{ISAC} has emerged as a promising technology to address these challenges. By sharing hardware platforms and spectral resources, an ISAC-enabled base station (BS) can simultaneously perform wireless sensing for vehicle state estimation and communication for C\&C signal transmission. This integration improves spectrum efficiency, reduces signaling latency, and strengthens the coordination between sensing and communication. Owing to these advantages, ISAC has attracted increasing attention in vehicular networks. Considering the high mobility of vehicles and the resulting rapidly time-varying wireless channels, the majority of existing works focus on predictive beamforming. In particular, an extended Kalman filter (EKF)-based predictive beamforming framework was first proposed in \cite{Liu}, enabling accurate vehicle tracking and reliable downlink communication without relying on downlink pilots or uplink feedback. Building upon this framework, the works in \cite{Du, extended_target} further modeled vehicles as extended targets rather than point-like objects, where more flexible beamforming strategies were designed to generate wide sensing beams and improve tracking accuracy. To handle dynamic vehicle behaviors and reduce tracking complexity, deep learning-based predictive beamforming methods were developed in \cite{DL3, DL2, DL1}, showing improved adaptability in complex vehicular environments.

Nevertheless, existing ISAC-enabled vehicular networks mainly rely on centralized sensing, where vehicle states are sensed and tracked by a single BS. Such a design may be inadequate for intersection management due to limited sensing coverage and severe blockage or occlusion caused by vehicles and surrounding buildings. These limitations motivate distributed and cooperative ISAC architectures for more robust vehicular sensing and coordination. Existing studies on distributed ISAC have primarily focused on cellular or cell-free networks with relatively static users. Based on the cooperation mode among BSs, these works can be broadly classified into three categories: single-active-BS architectures, where one BS transmits sensing signals and multiple BSs passively receive echoes \cite{one_bs_passive_sensing1, one_bs_passive_sensing2}; multi-active-BS architectures, where multiple BSs actively transmit sensing signals and cooperatively receive echoes \cite{passive_sensing1, passive_sensing2, passive_sensing3}; and fully cooperative architectures, where all BSs jointly participate in sensing transmission, echo reception, and signal processing \cite{all_active_sensing1, all_active_sensing2, all_active_sensing3}. However, methods designed for static users cannot be directly applied to vehicle coordination at intersections. Intersection management is a long-term task, where sensing and C\&C signals should be evaluated not only by their immediate effects, but also by their contributions to future trajectories, collision avoidance, and traffic efficiency. Existing distributed ISAC designs typically optimize sensing and communication performance independently in each time slot, overlooking temporal coupling among decisions and the task-level relevance of transmitted signals. Consequently, ISAC signals may be transmitted in every time slot regardless of their actual contribution to the coordination objective, causing redundant transmissions and inefficient resource utilization.

Goal-oriented semantic communication (GSC) \cite{GOSC} has recently emerged as a promising paradigm to address redundant data transmission by extracting and transmitting only the semantic representation that directly contributes to the task at the receiver. Unlike deep joint source-channel coding (DeepJSCC) \cite{JSCC}, which relies on end-to-end deep neural network training and defines semantics via latent feature embeddings, GSC derives interpretable semantic representations from raw data according to its interpretable relationship with the task objective. Also, GSC operates at both semantic and effectiveness levels, enabling communication decisions to be made based on the long-term task contribution \cite{SR1, SR2, SR3}. 
Owing to these advantages, recent studies have applied GSC to C\&C transmission in robotic systems, where the semantic representation of C\&C data corresponds to the critical control information required for motion control, such as thrust, roll angle, yaw angle, and velocity \cite{xu}. Specifically, \cite{xu} defined semantic- and effectiveness-level metrics for C\&C signals, enabling the BS to selectively transmit task-critical C\&C signals. Extended from \cite{xu}, \cite{wu} ranked queued C\&C semantic representations according to their age of information (AoI) and value of information (VoI), thereby assigning higher processing priority to more task-critical C\&C signals.
However, both studies considered point-to-point robotic control and assumed accurate real-time robot state information at the BS. These assumptions make them difficult to extend to practical intersection management, where vehicle states are affected by sensing uncertainty and control decisions are tightly coupled with the trajectory evolution of multiple interacting vehicles.

To fill this gap, we consider distributed ISAC-enabled vehicle coordination at intersections. In this scenario, multiple roadside units (RSUs) are deployed along different roads of an intersection and are responsible for transmitting sensing signals for vehicle state acquisition and C\&C signals for vehicle movement control. A central BS acts as the intersection manager, which collects sensing results from distributed RSUs, performs data fusion, and generates corresponding C\&C signals. Our goal is to enable efficient RSU signaling from a GSC perspective, where sensing and C\&C signals are transmitted only when they are semantically important for improving intersection traffic throughput. The main contributions of this paper are summarized as follows.
\begin{itemize}
    \item We consider distributed ISAC-enabled vehicle coordination at intersections, where multiple RSUs collaboratively perform vehicle sensing and C\&C signal transmission. To enable efficient signaling from a GSC perspective, our objective is twofold: maximizing intersection traffic throughput while minimizing the total number of transmitted sensing and C\&C signals.

    \item To achieve the above objective, we propose a novel unified GSC framework that tightly couples vehicle state prediction, VoI-based transmission decision-making, and uncertainty-aware wireless transmission design in a closed-loop manner. Specifically, the framework consists of three interdependent components:
    \begin{enumerate}
        \item We adopt an EKF at the BS to continuously predict vehicle states (e.g., speed and position), enabling the BS to obtain reference vehicle state information without the need to transmit sensing signals at every time slot. When sensing signals are transmitted by distributed RSUs, the EKF can also fuse the sensing measurements with the prediction results to correct estimation errors and reduce state uncertainty.

        \item Based on the estimated vehicle states and uncertainty covariances, we develop a masked hybrid proximal policy optimization (MHPPO) framework at the BS to jointly determine sensing transmission decisions, C\&C transmission decisions, and C\&C signal contents for different RSUs and vehicles. Guided by a VoI-based reward, MHPPO learns to trigger sensing and C\&C transmissions only when they are semantically important for improving traffic throughput.

        \item Given the decisions made by MHPPO, we propose an uncertainty-aware transmission design (UTD) for each RSU, including robust beamforming and VoI-based time-division power allocation. Robust beamforming incorporates vehicle state covariance information to maintain beam coverage over uncertain vehicle locations, while VoI-based time-division power allocation prioritizes effectiveness-critical C\&C transmissions and mitigates inter-RSU interference to enhance communication reliability.
    \end{enumerate}
    
    \item Extensive simulation results validate the effectiveness of our proposed GSC framework. Compared with predictive ISAC baselines adapted from state-of-the-art related studies and several ablation baselines, GSC achieves 100\% collision-free vehicle coordination with significantly reduced signaling overhead. Our results further confirm that EKF-based state prediction, MHPPO-based VoI-driven scheduling, and UTD-based transmission are jointly essential to achieve safe and signaling-efficient vehicle coordination.
\end{itemize}

\textit{Notations:} Throughout this paper, boldface lowercase letters, e.g., $\mathbf{a}$, and boldface uppercase letters, e.g., $\mathbf{A}$, are used to denote vectors and matrices, respectively. Constants are denoted by uppercase letters, e.g., A. 
$\mathbf{0}_{\mathrm{N}}$ and $\mathbf{I}_{\mathrm{N}}$ denote the $\mathrm{N}$-dimensional zero vector and the $\mathrm{N}\times\mathrm{N}$ identity matrix, respectively.
$\|\cdot\|_2$ denotes the Euclidean norm. 
The transpose, conjugate, Hermitian transpose, inverse, and trace of a matrix are denoted by $(\cdot)^\top$, $(\cdot)^*$, $(\cdot)^{\mathrm{H}}$, $(\cdot)^{-1}$, and $\mathrm{Tr}(\cdot)$, respectively. 
$\mathbb{C}^{\mathrm{M}\times\mathrm{N}}$ and $\mathbb{R}^{\mathrm{M}\times\mathrm{N}}$ denote the sets of $\mathrm{M}\times\mathrm{N}$ complex-valued and real-valued matrices, respectively. 
The Gaussian distribution and complex Gaussian distribution are denoted by $\mathcal{N}$ and $\mathcal{CN}$, respectively.


\section{System Model and Problem Formulation} \label{model}
In this section, we first introduce the distributed ISAC-enabled vehicle coordination task at an unsignalized intersection. We then present the sensing and C\&C signal transmission and reception models, followed by the kinematic vehicle model. Finally, we formulate an optimization problem that aims to improve intersection traffic throughput while minimizing the total number of transmitted sensing and C\&C signals.

\subsection{Scenario description}
\begin{figure}
    \centering
    \includegraphics[width=0.45\textwidth]{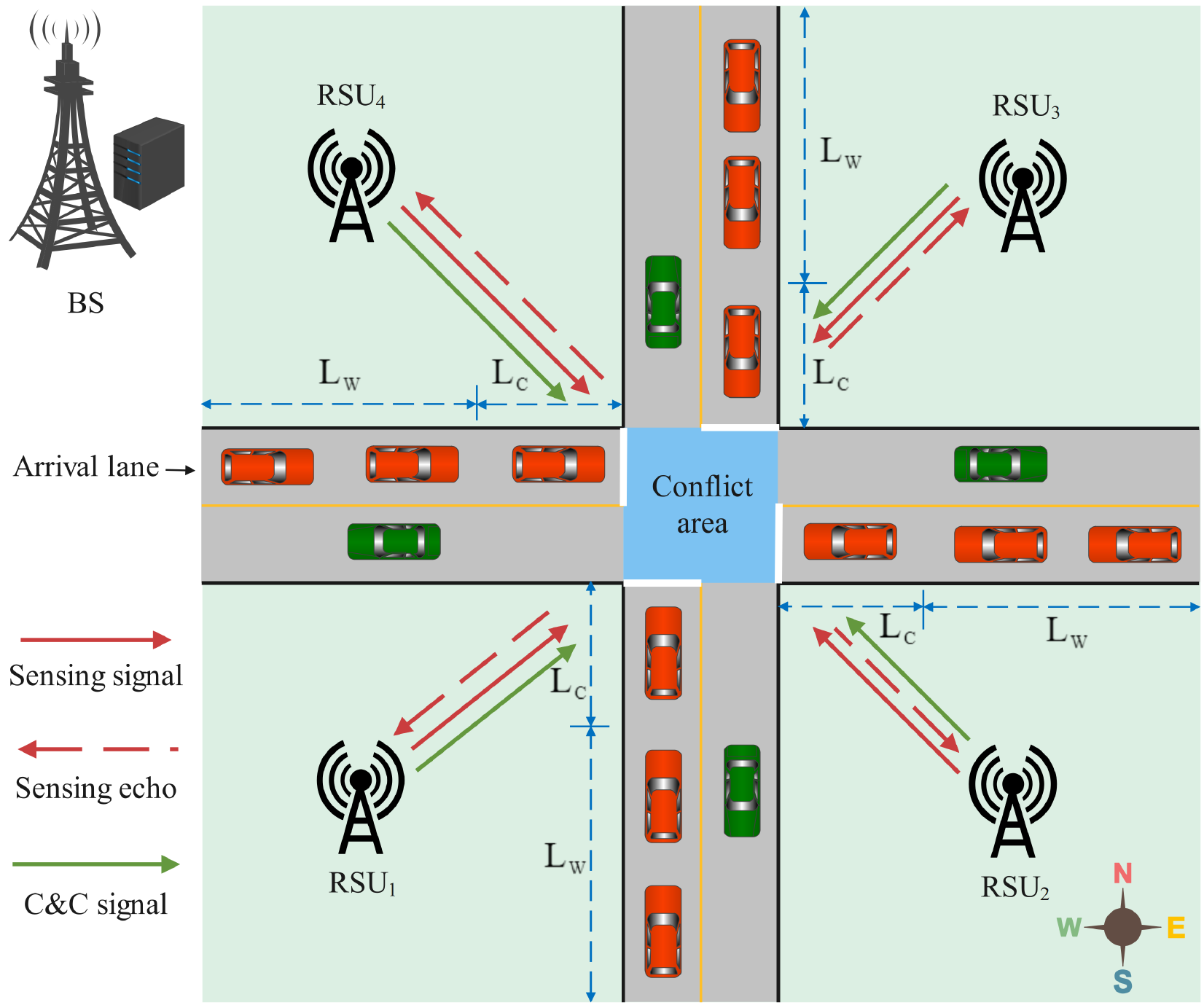}
    \caption{Illustration for the distributed ISAC-enabled automated vehicle coordination at an unsignalized intersection.}
    \label{system_model}
\end{figure}
As illustrated in \figref{system_model}, we consider a distributed ISAC-enabled vehicle coordination task at an unsignalized intersection. The intersection consists of four roads, each with a single arrival lane. The set of roads is denoted by $\mathcal{K}=\{1,\ldots,k,\ldots,\mathrm{K}\}$. For simplicity, the four roads are assumed to be aligned with the cardinal directions, i.e., north, east, south, and west. Along each road, a RSU is deployed to transmit both sensing and C\&C signals. Let $\mathcal{N} = \{1,\ldots,n,\ldots,\mathrm{N}\}$ denote the set of RSUs, and let the position of RSU $n$ in the Cartesian coordinate system be $\mathbf{p}_n = [x_n,\, y_n]^\top$. Since each RSU is uniquely associated with one road, there exists a one-to-one mapping between road index $k$ and RSU index $n$. The four RSUs are connected to a central BS via fiber-optic links. The BS is responsible for collecting sensing information from all RSUs, performing data fusion, and generating the corresponding C\&C signals. 

The intersection is divided into three regions: a waiting area with side length $\mathrm{L_w}$, a control area with side length $\mathrm{L_c}$, and a conflict area. In the waiting area, vehicles report their driving intentions, i.e., going straight, turning left, or turning right, to the RSUs and wait for coordination. Vehicles are not permitted to enter the control area until the preceding vehicle in the same lane has completely traversed the conflict area. The control area of each road can accommodate at most one vehicle at a time. We use $m_k$ to denote the $m$-th vehicle entering the control area of the $k$-th road. Once vehicle $m_k$ is granted access to the control area, it moves at a low speed and transfers its control authority to the central BS. Subsequently, the RSUs transmit sensing signals to estimate the vehicle states, such as position and velocity. The sensing results are sent back to the BS, where data fusion is performed to generate C\&C signals that determine vehicle accelerations. These C\&C signals are then forwarded to the RSUs and transmitted to the corresponding vehicles. This sensing-control loop is repeated until the vehicle successfully passes through the conflict area.

Our objective is to maximize the traffic throughput of the unsignalized intersection while minimizing the number of transmitted sensing and C\&C signals within a given time horizon $\mathrm{T}$. For tractability, $\mathrm{T}$ is discretized into $\mathrm{I}$ time slots, indexed by $t_i$, where $i \in \{0,1,2,\ldots,\mathrm{I}-1\}$. The time slot duration is given by $\Delta t = t_{i+1}-t_i$, $\forall i$, where $\Delta t$ is constant and satisfies $\Delta t=\mathrm{T}/\mathrm{I}$. Let $\mathcal{M}_{k}(t_i)$ denote the set of vehicles that enter the intersection from road $k$ and have successfully exited the conflict area by time $t_i$.

Each RSU is equipped with $\mathrm{N_t}$ transmit antennas and $\mathrm{N_r}$ receive antennas arranged as uniform linear arrays (ULAs), which are used to transmit millimeter-wave (mmWave) orthogonal frequency-division multiplexing (OFDM) signals for sensing and communication. Specifically, for sensing, the $n$-th RSU monitors all vehicles within the conflict area and the control areas of the intersection. For communication, it serves only its associated vehicle $m_k$. For simplicity, we assume that the transmit and receive arrays are physically separated with sufficient isolation, such that the self-interference introduced by full-duplex operation is negligible~\cite{self-interference}. The antenna element spacing of each ULA is $\mathrm{D}=\lambda_c/2$, where $\lambda_c$ is the carrier wavelength.
The angular scanning range of each ULA is $[-90^\circ,90^\circ]$, and the ULA of each RSU is oriented toward its corresponding road. For example, under the left-lane driving convention, e.g., the U.K. traffic regulation, the ULA of the RSU deployed on the west road is oriented toward the south.

\subsection{Transmit Signal Model}
Let $\mathrm{N_s}$ and $\mathrm{N_c}$ denote the numbers of sensing and communication subcarriers, respectively, and let $\Delta f$ denote the subcarrier spacing. Sensing and communication are assumed to operate over separate frequency bands for two main reasons. First, sensing typically requires a large bandwidth, e.g., more than $100$ MHz, to achieve high-precision estimation, whereas C\&C signal transmission may only occupy a much narrower bandwidth, e.g., less than $1$ MHz. Second, since sensing and C\&C signals are directed toward the same vehicles, using separate frequency bands facilitates reliable signal separation at the vehicles.
Let $\mathrm{N}^s_{\mathrm{sym}}$ and $\mathrm{N}^c_{\mathrm{sym}}$ denote the numbers of sensing and communication OFDM symbols within one time slot, respectively. The duration of each OFDM symbol is given by $\mathrm{T_{sym}}=1/\Delta f + \mathrm{T_{cp}}$, where $\mathrm{T_{cp}}$ is the cyclic prefix duration. The transmit signal of the $n$-th RSU in the $t_i$-th time slot is expressed as
\begin{equation}  \label{transmit_signal}
\begin{aligned}
\mathbf{x}_n(t_i)
=&~ \delta_n^s(t_i)\sqrt{p^s_n(t_i)}\,\mathbf{w}_n^s(t_i)\,\mathrm{x}_n^s(t_i) \\
&+ \delta_n^c(t_i)\sqrt{p^c_n(t_i)}\,\mathbf{w}_n^c(t_i)\,\mathrm{x}_n^c(t_i),
\end{aligned}
\end{equation}
where $\delta_n^s(t_i), \delta_n^c(t_i) \in \{0,1\}$ are binary indicators specifying whether the $n$-th RSU transmits sensing and C\&C signals, respectively. We assume that the RSU can transmit a C\&C signal at time $t_{i+1}$ only if it performed sensing at time $t_{i}$. This assumption is motivated by the fact that mmWave communication highly depends on accurate directional beamforming enabled by prior sensing. The terms $p^s_n(t_i)$ and $p^c_n(t_i)$ denote the transmit power for sensing and communication, respectively; $\mathbf{w}_n^s(t_i), \mathbf{w}_n^c(t_i) \in \mathbb{C}^{\mathrm{N_t} \times 1}$ are the corresponding normalized beamforming vectors satisfying $\|\mathbf{w}_n^s(t_i)\|_2 = 1$ and $\|\mathbf{w}_n^c(t_i)\|_2 = 1$, while $\mathrm{x}_n^s(t_i)$ and $\mathrm{x}_n^c(t_i)$ denote the transmitted sensing and communication symbols, respectively.

In general, coherent distributed sensing requires strict time and phase synchronization among RSUs, which leads to significant implementation complexity~\cite{non-coherent}. Therefore, in this work, we adopt non-coherent distributed sensing due to its practical advantages. To mitigate mutual interference among RSUs under non-coherent operation, orthogonal coding, e.g., Walsh--Hadamard coding, is employed for sensing signals. Specifically, for each time slot $t_i$ consisting of $\mathrm{N}^s_{\mathrm{sym}}$ OFDM symbols, the sensing signals are designed to satisfy
\begin{equation}
\sum_{n_1=0}^{\mathrm{N}^s_{\mathrm{sym}}-1} 
\sum_{n_2=0}^{\mathrm{N_s}-1} 
\mathrm{x}_n^s[n_1,n_2]
\big(\mathrm{x}_{\widetilde{n}}^s[n_1,n_2]\big)^*
= 0, 
\quad \forall \widetilde{n} \in \mathcal{N},~ \widetilde{n} \neq n,
\end{equation}
where $\mathrm{x}_{n}^s[n_1,n_2]$ denotes the $n_2$-th subcarrier sample of the $n_1$-th OFDM symbol transmitted by the $n$-th RSU. For notational simplicity, the time index $t_i$ is omitted in the following discussion unless explicitly specified.

\vspace{-1pt}
\subsection{Sensing Model}

The transmitted sensing signals are reflected by vehicles located in the control and conflict areas and are subsequently collected by each RSU. Specifically, the received sensing echo corresponding to the $n_1$-th OFDM symbol at the $n$-th RSU is expressed in \eqref{sensing_rx}.
\begin{figure*}[!t]
\begin{equation} \label{sensing_rx}
\begin{aligned}
\mathbf y_n(t_i,n_1) =&\, \sum_{k=1}^{\mathrm{K}}\delta_{n}^s \sqrt{\mathrm{N_t}\mathrm{N_r}} \sqrt{p^s_n}  \beta_{n}^{m_k}e^{j2\pi f_n^{m_k}n_1\mathrm{T_{sym}}}\mathbf{a}_\mathrm{r}(\theta_{n}^{m_k}) \mathbf{a}_\mathrm{t}^{\mathrm{H}}(\theta_{n}^{m_k})\mathbf w_n^s \mathrm{x}_n^s(t_i-2\tau_n^{m_k}) + \sum_{\substack{\widetilde{n}\in\mathcal{N} \\ \widetilde{n}\neq n}}\sum\limits_{k=1}^{\mathrm{K}}\delta_{\widetilde{n}}^s \sqrt{\mathrm{N_t}\mathrm{N_r}} \\
& \cdot \sqrt{p_{\widetilde{n}}^s}\beta_{n,\widetilde{n}}^{m_k} e^{j2\pi f_{n,\widetilde{n}}^{m_k} n_1\mathrm{T_{sym}}}\mathbf{a}_\mathrm{r}(\theta_{n}^{m_k})\mathbf{a}_\mathrm{t}^{\mathrm{H}}(\theta_{\widetilde{n}}^{m_k})\mathbf w_{\widetilde{n}}^s \mathrm{x}_{\widetilde{n}}^s(t_i-\tau_{n}^{m_k}-\tau_{\widetilde{n}}^{m_k})  +  \mathbf{z}^{\text{clt}} + \mathbf{z}^s,
\end{aligned}
\end{equation}
\hrulefill 
\vspace*{-6pt} 
\end{figure*}
The first term of on the right-hand side (RHS) of \eqref{sensing_rx} represents the monostatic sensing echo: 
$\beta_n^{m_k} = \sqrt{\frac{\varepsilon_n^{m_k}\lambda_c^2}{(4\pi)^3(d_n^{m_k})^4}}$ represents the attenuation coefficient between RSU $n$ and vehicle $m_k$, in which $\varepsilon_n^{m_k}$ is the radar cross section (RCS) of vehicle $m_k$ from the perspective of RSU $n$, $d_n^{m_k}$ is the Euclidean distance between RSU $n$ and vehicle $m_k$;
$f_n^{m_k} = v_{m_k}\cos(\phi_n^{m_k})/\lambda_c$ is the Doppler shift, in which $v_{m_k}$ is the speed of vehicle $m_k$, $\phi_n^{m_k}$ is the angle between the vehicle's direction of motion and the line-of-sight (LoS) from vehicle $m_k$ to RSU $n$;
$2\tau_{n}^{m_k}=2d_n^{m_k}/c$ is the round-trip delay, $c$ is the speed of light.
The second term on the RHS represents the bistatic sensing echo: $\beta_{n,\widetilde{n}}^{m_k} = \sqrt{\frac{\varepsilon_{n,\widetilde{n}}^{m_k}\lambda_c^2}{(4\pi)^3(d_n^{m_k})^2(d_{\widetilde{n}}^{m_k})^2}}$, where $\varepsilon_{n,\widetilde{n}}^{m_k}$ denotes the bistatic RCS of vehicle $m_k$ with respect to the transmitter--receiver pair $(\widetilde{n}, n)$; $f_{n,\widetilde{n}}^{m_k}
= v_{m_k}(\cos(\phi_n^{m_k})+\cos(\phi_{\widetilde{n}}^{m_k}))/\lambda_c$
denotes the bistatic Doppler shift. 
The third term on the RHS $\mathbf{z}^{\text{clt}}\in \mathbb{C}^{\mathrm{N_r}\times 1}$ is the clutter noise, and $\mathbf{z}^{\text{clt}} \sim \mathcal{CN}(\mathbf{0}_{\mathrm{N_r}}, \mathbf{R}^{\text{clt}})$, where $\mathbf{R}^{\text{clt}} \in \mathbb{C}^{\mathrm{N_r}\times \mathrm{N_r}}$ is the clutter covariance matrix and is denoted as 
$\mathbf{R}^{\text{clt}} = \sum_{q=1}^{Q} \sigma_{q}^2 \mathbf{a}_{\text{r}}(\theta_q^{\text{clt}}) \mathbf{a}_{\text{r}}^{\mathrm{H}}(\theta_q^{\text{clt}})$ \cite{clutter}, where $Q$ is the total number of stationary scatterers, $\sigma_{q}^2$ is the average power of the $q$-th scatterer, and $\mathbf{a}_{\text{r}}(\theta_q^{\text{clt}})$ is the receive steering vector for the clutter angle.
The fourth term on the RHS $\mathbf{z}^s \in \mathbb{C}^{\mathrm{N_r}\times 1}$ denotes complex additive white Gaussian noise (AWGN) with zero mean and variance of $\sigma^2_s=$, i.e., $\mathbf{z}^s\sim \mathcal{CN}(\mathbf{0}_{\mathrm{N_r}}, \sigma^2_s\mathbf{I}_{\mathrm{N_r}})$, $\sigma_s^2=\mathrm{N_s}\Delta f\sigma^2_0$ and $\sigma^2_0$ is the  power spectral density of AWGN.
The receive and transmit steering vectors $\mathbf{a}_\mathrm{r}(\cdot)\in\mathbb{C}^{\mathrm{N_r}\times 1}$ and $\mathbf{a}_\mathrm{t}(\cdot)\in\mathbb{C}^{\mathrm{N_t}\times 1}$ are respectively defined as
\begin{align}
    \mathbf{a}_\mathrm{r}(\cdot)
    &=\frac{1}{\sqrt{\mathrm{N_r}}} \bigg[1, e^{-j2\pi\frac{\mathrm{D}\sin(\cdot)}{\lambda_c}}, \ldots, e^{-j2\pi\frac{(\mathrm{N_r}-1)\mathrm{D}\sin(\cdot)}{\lambda_c}}\bigg]^\top, \\
    \mathbf{a}_\mathrm{t}(\cdot)
    &= \frac{1}{\sqrt{\mathrm{N_t}}}\bigg[1, e^{-j2\pi\frac{\mathrm{D}\sin(\cdot)}{\lambda_c}}, \ldots, e^{-j2\pi\frac{(\mathrm{N_t}-1)\mathrm{D}\sin(\cdot)}{\lambda_c}}\bigg]^\top.
\end{align}

After matched filtering, the delay and Doppler shift of vehicle $m_k$ observed at the $n$-th RSU are estimated as
\begin{equation}
    \hat{\tau}_n^{m_k}= \tau_n^{m_k} + z_n^{m_k}(\tau), \quad \hat{f}_n^{m_k} = f_n^{m_k} + z_n^{m_k}(f),
\end{equation}
where $z_n^{m_k}(\tau)$ and $z_n^{m_k}(f)$ denote zero-mean measurement errors with variances of $\sigma^2_{n,m_k}(\tau)$ and $\sigma^2_{n,m_k}(f)$, respectively. According to \cite{Liu}, the variances $\sigma^2_{n,m_k}(\tau)$ and $\sigma^2_{n,m_k}(f)$ are inversely proportional to the signal-to-interference-plus-noise ratio (SINR) corresponding to \eqref{sensing_rx}. The corresponding proportional relationship can be expressed as \eqref{sensing_sigma_3}, shown at the top of the next page due to space limitations, 
\begin{figure*}[!t]
\begin{equation} \label{sensing_sigma_3}
\sigma_{n,m_k}^2(\tau), \sigma_{n,m_k}^2(f)
\propto 
\frac{
\sum\limits_{\substack{\widetilde{n}\in\mathcal{N} \\ \widetilde{n}\neq n}}\sum\limits_{\ell=1}^{\mathrm{K}}\delta_{\widetilde{n}}^s\mathrm{N_t}\mathrm{N_r} p_{\widetilde{n}}^s|\rho_{n,\widetilde{n}}|^2|\beta_{n,\widetilde{n}}^{m_\ell}|^2 | \mathbf{a}_\mathrm{t}^{\mathrm{H}}(\theta_{\widetilde{n}}^{m_\ell})\mathbf w_{\widetilde{n}}^s|^2+ \sum\limits_{q=1}^Q\sigma^2_q + \sigma_s^2
}
{
\eta \mathrm{N_t}\mathrm{N_r} p_n^s|\beta_n^{m_k}|^2 |\mathbf{a}_\mathrm{t}^{\mathrm{H}}(\theta_{n}^{m_k})\mathbf w_{n}^s|^2
},
\end{equation}
\hrulefill 
\end{figure*}
where $\rho_{n,\widetilde{n}}$ denotes the cross-correlation coefficient between the sensing waveforms of RSUs $n$ and $\widetilde{n}$, satisfying $0\leq|\rho_{n,\widetilde{n}}|\ll 1$ due to orthogonal coding, and $\eta=\mathrm{N}^s_{\mathrm{sym}}$ denotes the matched-filtering gain. Since orthogonal coding makes the residual inter-RSU sensing interference negligible, the estimation error variances are approximated as
\begin{align}
    &\sigma^2_{n,m_k}(\tau) = \frac{\alpha_1^2\Big(\sigma^2_s + \sum\limits_{q=1}^Q\sigma^2_q\Big)}{\eta \mathrm{N_t}\mathrm{N_r} p_n^s|\beta_n^{m_k}|^2 |\mathbf{a}_\mathrm{t}^{\mathrm{H}}(\theta_{n}^{m_k})\mathbf w_{n}^s|^2}, \\
    &\sigma^2_{n,m_k}(f) = \frac{\alpha_2^2\Big(\sigma^2_s + \sum\limits_{q=1}^Q\sigma^2_q\Big)}{\eta \mathrm{N_t}\mathrm{N_r} p_n^s|\beta_n^{m_k}|^2 | \mathbf{a}_\mathrm{t}^{\mathrm{H}}(\theta_{n}^{m_k})\mathbf w_{n}^s|^2},
\end{align}
where $\alpha_1$ and $\alpha_2$ are constants related to the system configuration, signal design, and the adopted signal processing algorithms \cite{Du}. After separating vehicle echoes through matched filtering, the angle of arrival (AoA) of each vehicle $\theta_n^{m_k}$ can be estimated using maximum likelihood estimation
(MLE) or multiple signal classification (MUSIC). The estimated AoA is given by
\begin{equation}
    \hat{\theta}_n^{m_k} = \theta_n^{m_k} + z_n^{m_k}(\theta),
\end{equation}
where $z_n^{m_k}(\theta)$ denotes the zero-mean AoA measurement error with variance $\sigma^2_{n,m_k}(\theta)$. According to \cite{Du}, the AoA estimation variance $\sigma^2_{n,m_k}(\theta)$ can be approximated as
\begin{equation}
    \sigma^2_{n,m_k}(\theta) = \frac{\alpha_3^2\Big(\sigma^2_s + \sum\limits_{q=1}^Q\sigma^2_q\Big)}{\eta \mathrm{N_t}\mathrm{N_r}p_n^s|\beta_n^{m_k}|^2 |\mathbf{a}_\mathrm{t}^{\mathrm{H}}(\theta_{n}^{m_k})\mathbf w_{n}^s|^2}, 
\end{equation}
in which $\alpha_3$ is a constant related to the system configuration, signal design, and the adopted signal processing algorithms \cite{Du}. 
The sensing results obtained from multiple RSUs are then fused at the central BS. Since the delay, Doppler shift, and AoA measurements are obtained in the local coordinate systems of different RSUs, they cannot be directly fused. To address this issue, we first transform the local sensing measurements into the Cartesian domain and then fuse them in the global Cartesian coordinate system. The global coordinate system is defined such that the positive $y$-axis points north and the positive $x$-axis points east. Let the local Cartesian measurement of vehicle $m_k$ obtained from RSU $n$ be denoted by
$\hat{\mathbf{s}}_n^{m_k} = [\hat{x}_n^{m_k},\, \hat{y}_n^{m_k},\, \hat{v}_n^{m_k}]^\top$, and let the fused global measurement at the BS be denoted by
$\hat{\mathbf{s}}_{m_k}^G = [\hat{x}_{m_k}^G,\, \hat{y}_{m_k}^G,\, \hat{v}_{m_k}^G]^\top$.
Specifically, the range and speed estimates obtained from RSU $n$ are given by
\begin{equation}
    \hat{r}_n^{m_k} = c \hat{\tau}_n^{m_k}, 
\quad
    \hat{v}_n^{m_k} = \frac{\lambda_c \hat{f}_n^{m_k}}{\cos(\phi_n^{m_k})}.
\end{equation}

Since the AoA is measured clockwise from the north direction, the local Cartesian measurement can be written as
\begin{equation}
    \hat{\mathbf{s}}_n^{m_k} =
\begin{bmatrix}
x_n + \hat{r}_n^{m_k}\sin(\hat{\theta}_n^{m_k}) \\
y_n + \hat{r}_n^{m_k}\cos(\hat{\theta}_n^{m_k}) \\
\hat{v}_n^{m_k}
\end{bmatrix}.
\end{equation}

Let $\hat{\mathbf{R}}_n^{m_k} = \mathrm{diag}\big(\sigma^2_{n,m_k}(\tau),\, \sigma^2_{n,m_k}(f),\, \sigma^2_{n,m_k}(\theta)\big)$ denote the covariance matrix of the raw sensing measurements of vehicle $m_k$ at RSU $n$. By applying first-order error propagation, the covariance matrix of $\hat{\mathbf{s}}_n^{m_k}$ is obtained as
\begin{equation}
    \hat{\mathbf{P}}_n^{m_k} = \mathbf{J}_n^{m_k} \hat{\mathbf{R}}_n^{m_k} (\mathbf{J}_n^{m_k})^\top,
\end{equation}
where $\mathbf{J}_n^{m_k}$ is the Jacobian matrix defined as
\begin{equation}
\mathbf{J}_n^{m_k} = \begin{bmatrix}
c\sin(\hat{\theta}_n^{m_k}) & 0 & \hat{r}_n^{m_k}\cos(\hat{\theta}_n^{m_k}) \\
c\cos(\hat{\theta}_n^{m_k}) & 0 & -\hat{r}_n^{m_k}\sin(\hat{\theta}_n^{m_k}) \\
0 & \frac{\lambda_c}{\cos(\phi_n^{m_k})} & 0
\end{bmatrix}.
\end{equation}

Assuming that the BS employs optimal linear fusion, the global measurement and its corresponding variance for vehicle $m_k$ are given by
\begin{align}
\hat{\mathbf{s}}_{m_k}^G
&= \hat{\mathbf{P}}_{m_k}^G
\sum_{n\in\mathcal{N}} \delta_n^s (\hat{\mathbf{P}}_n^{m_k})^{-1} \hat{\mathbf{s}}_n^{m_k}, 
\quad \exists\, \delta_n^s = 1, \label{mea_fusion_value}  \\
\hat{\mathbf{P}}_{m_k}^G
&= \left( \sum_{n\in\mathcal{N}} \delta_n^s (\hat{\mathbf{P}}_n^{m_k})^{-1} \right)^{-1},
\quad \exists\, \delta_n^s = 1. \label{mea_fusion_var} 
\end{align}


\subsection{Communication Model}
Each vehicle is equipped with a single receive antenna. As previously stated, vehicle $m_k$ is served by its associated RSU $n$. However, in the conflict area, vehicles may be located in close proximity. As a result, vehicle $m_k$ may also experience interference from other RSUs. Let $\mathcal{M}^{\text{com}}_n$ denote the set of vehicles within the communication coverage of RSU $n$. 
The received communication signal at vehicle $m_k$ is given by
\begin{equation} \label{com_rx}
\begin{aligned}
    \mathrm y_{m_k}=\,& \delta_n^c \sqrt{\mathrm{N_t}}\sqrt{p_n^c}\kappa_n^{m_k}\mathbf{a}_\mathrm{t}^{\mathrm{H}}(\theta_{n}^{m_k})\mathbf w_n^c\mathrm{x}_n^c(t_i-\tau_n^{m_k}) \\
     &+ 
     \sum_{\substack{\widetilde{n}: m_k\in\mathcal{M}_{\widetilde{n}}^{\text{com}} \\ \widetilde{n}\neq n}} \delta_{\widetilde{n}}^c \sqrt{\mathrm{N_t}}\sqrt{p_{\widetilde{n}}^c}\kappa_{\widetilde{n}}^{m_k}\mathbf{a}_\mathrm{t}^{\mathrm{H}}(\theta_{\widetilde{n}}^{m_k})\mathbf w_{\widetilde{n}}^c \\
     &\cdot \mathrm{x}_{\widetilde{n}}^c(t_i-\tau_{\widetilde{n}}^{m_k}) + \mathrm{z}^{\text{clt}} +  \mathrm z^c, 
\end{aligned}
\end{equation}
where $\kappa_n^{m_k} = \frac{\lambda_c}{4\pi d_n^{m_k}} e^{-j2\pi d_n^{m_k}/\lambda_c}$ represents the LoS channel coefficient; $\mathrm{z}^{\text{clt}} \sim \mathcal{CN}(0, \sigma_{\text{clt}}^2)$ denotes the multipath clutter noise; $\mathrm z^c \sim \mathcal{CN}(0, \sigma_c^2)$ denotes AWGN with variance $\sigma_c^2$, where $\sigma_c^2=\mathrm{N_c}\Delta f\sigma^2_0$. Accordingly, the received SINR at vehicle $m_k$ is expressed as 
\begin{equation}
    \mathrm{SINR}_{m_k} = \frac{\delta_n^c\mathrm{N_t}p_n^c |\kappa_n^{m_k}|^2 |\mathbf{a}_\mathrm{t}^{\mathrm{H}}(\theta_{n}^{m_k})\mathbf{w}_n^c|^2}{ I_{m_k} + \sigma_{\text{clt}}^2 + \sigma^2_c },
\end{equation}
where $I_{m_k}$ denotes the aggregate inter-cell interference power and is defined as
\begin{equation}
    I_{m_k} = \sum_{\substack{\widetilde{n}: m_k\in\mathcal{M}_{\widetilde{n}}^{\text{com}} \\ \widetilde{n}\neq n}}\delta_{\widetilde{n}}^c\mathrm{N_t}p_{\widetilde{n}}^c|\kappa_{\widetilde{n}}^{m_k}|^2 |\mathbf{a}_\mathrm{t}^{\mathrm{H}}(\theta_{\widetilde{n}}^{m_k})\mathbf w_{\widetilde{n}}^c|^2.
\end{equation}

To guarantee reliable decoding of the C\&C signal, the SINR must satisfy the following constraint
\begin{equation} 
\mathrm{SINR}_{m_k} \geq \gamma_{\mathrm{thr}}, \quad \text{if } \delta_n^c=1.
\end{equation}
where $\gamma_{\mathrm{thr}}$ denotes the SINR threshold required for successful decoding.

\subsection{Vehicle Kinematic Model}
To capture the stochastic nature of traffic dynamics, each vehicle exhibits a random maneuver intention, i.e., going straight, turning left, or turning right. Owing to lane geometry or traffic regulations, the trajectory corresponding to each maneuver is assumed predefined. 
The kinematic bicycle model \cite{bicycle_model} is adopted to characterize the vehicle motion.
The state vector of vehicle $m_k$ in the Cartesian coordinate system is defined as $\mathbf{s}_{m_k}= [x_{m_k},\, y_{m_k},\, \psi_{m_k},\, v_{m_k}]^\top$, where $[x_{m_k},\, y_{m_k}]^\top$ denotes the position, and $\psi_{m_k}$ denotes the heading angle. The vehicle dynamics are modeled as
\begin{equation}  \label{veh_dynamics}
\mathbf{s}_{m_k}(t_i) = f(\mathbf{s}_{m_k}(t_{i-1}),\, \mathbf{u}_{m_k}(t_{i-1})) + \mathbf{n}_{m_k}(t_i),
\end{equation}
where $\mathbf{u}_{m_k}(t_{i-1}) = [\omega_{m_k}(t_{i-1}),\, a_{m_k}(t_{i-1})]^\top$ is the control vector, $\omega_{m_k}(t_{i-1})$ is the steering angle, $a_{m_k}(t_{i-1})$ is the acceleration, and $\mathbf{n}_{m_k}(t_i)\in\mathbb{R}^{4\times1}$ denotes the process noise. Although the steering angle is used in the motion model, it is determined by the predefined route geometry and is not optimized by the BS. Consequently, the BS only needs to determine the vehicle acceleration. The state transition function $f(\mathbf{s}_{m_k}(t_{i-1}),\mathbf{u}_{m_k}(t_{i-1}))$ is defined as
\begin{equation} \label{bicycle_func}
\begin{aligned}
&f\big(\mathbf{s}_{m_k}(t_{i-1}),\mathbf{u}_{m_k}(t_{i-1})\big)  \\
&=
\begin{bmatrix} 
x_{m_k}(t_{i-1}) + v_{m_k}(t_{i-1}) \cos(\psi_{m_k}(t_{i-1})) \Delta t \\ 
y_{m_k}(t_{i-1}) + v_{m_k}(t_{i-1}) \sin(\psi_{m_k}(t_{i-1})) \Delta t \\ 
\psi_{m_k}(t_{i-1}) + \frac{v_{m_k}(t_{i-1})}{\mathrm{L_{wb}}} \tan(\omega_{m_k}(t_{i-1})) \Delta t \\ 
v_{m_k}(t_{i-1}) + a_{m_k}(t_{i-1}) \Delta t
\end{bmatrix},
\end{aligned}
\end{equation}
where $\mathrm{L_{wb}}$ represents the wheelbase, i.e., the distance between the front and rear axles.
Due to physical constraints of vehicle motion, the speed and acceleration are bounded as
\begin{equation} \label{speed_constraint}
0 \leq v_{m_k} \leq \mathrm{V}_{\max}, \quad
-\mathrm{A}_{\max} \leq a_{m_k} \leq \mathrm{A}_{\max},
\end{equation}
where $\mathrm{V}_{\max}$ is the maximum allowable speed and $\mathrm{A}_{\max}$ denotes the maximum acceleration. 
Moreover, to ensure practical collision avoidance, we represent the road region occupied by the vehicle as a convex set instead of a point. Let $\mathrm{L}$ and $\mathrm{W}$ denote the length and width of the vehicle, respectively. The occupied region corresponding to state $\mathbf{s}_{m_k}$ is defined as the following rectangular convex set \cite{convex_set}
\begin{align}
    &\mathcal{C}(\mathbf{s}_{m_k})=\Big\{\mathbf{c}\in \mathbb{R}^2 | \mathbf{A}(\mathbf{s}_{m_k})\mathbf{c} \leq \mathbf{b}(\mathbf{s}_{m_k}) \Big\}, \\
    & \mathbf{A}(\mathbf{s}_{m_k}) = \Big[\mathbf{A}_1(\psi_{m_k}),\, -\mathbf{A}_1(\psi_{m_k})^\top \Big]^\top, \\
    & \mathbf{b}(\mathbf{s}_{m_k}) = \Big[\frac{\mathrm{L}}{2},\frac{\mathrm{W}}{2},\frac{\mathrm{L}}{2},\frac{\mathrm{W}}{2} \Big]^\top + \mathbf{A}(\mathbf{s}_{m_k})\mathbf{p}_{m_k},
\end{align}
where $\mathbf{A}_1(\psi_{m_k}) = 
\begin{bmatrix}
\cos(\psi_{m_k}) & -\sin(\psi_{m_k}) \\
\sin(\psi_{m_k}) & \cos(\psi_{m_k})
\end{bmatrix}$ is the rotation matrix, and $\mathbf{p}_{m_k} = [x_{m_k}, y_{m_k}]^\top$ denotes the position of vehicle $m_k$. To ensure collision avoidance between any two distinct vehicles, the following condition should be satisfied
\begin{equation} \label{collision_avoid}
\mathcal{C}(\mathbf{s}_{m_k}) \cap \mathcal{C}(\mathbf{s}_{m_{\widetilde{k}}}) = \varnothing, 
\quad \forall\, k,\widetilde{k} \in \mathcal{K},~ \widetilde{k} \neq k.
\end{equation}

\subsection{Problem Formulation and Decomposition}
Our goal is twofold: 1) to maximize the traffic throughput at the unsignalized intersection, and 2) to minimize the number of transmitted sensing and C\&C signals. Accordingly, the optimization problems are formulated as
\begin{align}
    \mathcal{P}_{1.1}:\quad\mathop{\max}_{\{\mathcal{X}(t_i)\}_{i=0}^{\mathrm{I}-1}} &~ \mathbb{E}\Bigg[\sum_{k=1}^{\mathrm{K}} |\mathcal{M}_k(t_{\mathrm{I}-1})| \Bigg] \label{eq1} \\
    \textrm{s. t. } &~ \delta_n^s(t_i),\,\delta_n^c(t_i) \in\{0,\,1\},  \tag{\ref{eq1}{a}}   \label{eq1a} \\
    &~ \delta_n^c(t_{i+1}) \leq \delta_n^s(t_{i}),  \tag{\ref{eq1}{b}}   \label{eq1b} \\
    & ~ 0 \leq p^s_n(t_i),\, p^c_n(t_i) \leq \mathrm{P_{max}}, \tag{\ref{eq1}{c}}   \label{eq1c} \\
    &~  \delta_n^s(t_i)p^s_n(t_i) + \delta_n^c(t_i)p^c_n(t_i) \leq \mathrm{P_{max}}, \tag{\ref{eq1}{d}}   \label{eq1d} \\
    &~\|\mathbf{w}_n^s(t_i)\|^2_2=1,\,\|\mathbf{w}_n^c(t_i)\|^2_2 = 1, \tag{\ref{eq1}{e}}   \label{eq1e} \\
    &~ \mathrm{SINR}_{m_k}(t_i) \geq \gamma_{\mathrm{thr}},  \tag{\ref{eq1}{f}}   \label{eq1f} \\
    &~-\mathrm{A}_{\max} \leq a_{m_k}(t_i) \leq \mathrm{A}_{\max}, \tag{\ref{eq1}{g}}   \label{eq1g} \\
    &~ \mathcal{C}(\mathbf{s}_{m_k}(t_i)) \cap \mathcal{C}(\mathbf{s}_{m_{\widetilde{k}}}(t_i)) = \varnothing. \tag{\ref{eq1}{h}}   \label{eq1h} \\
    \mathcal{P}_{1.2}:\quad\mathop{\min}_{\{\mathcal{X}(t_i)\}_{i=0}^{\mathrm{I}-1}} &~ \mathbb{E}\Bigg[\sum_{i=0}^{\mathrm{I-1}}  \sum_{n=1}^{\mathrm{N}}(\delta_n^s(t_i)+\delta^c_n(t_i)) \Bigg] \label{eq1.2}  \\
    \textrm{s. t. } &~\eqref{eq1a}-\eqref{eq1h}, \notag 
\end{align}
where $\mathcal{X}(t_i) = \{ \delta_n^s(t_i), \delta_n^c(t_i), p^s_n(t_i), p^c_n(t_i), \mathbf{w}_n^s(t_i), \mathbf{w}_n^c(t_i),\\ a_{m_k}(t_i) \}$ denotes the set of optimization variables,
\eqref{eq1} aims to maximize the expected number of vehicles that successfully pass through the intersection over the considered time horizon; 
\eqref{eq1a} enforces binary decisions for sensing and C\&C signal transmissions, while \eqref{eq1b} ensures that a communication action can only occur if a sensing action has been performed in the previous time step; 
\eqref{eq1c} and \eqref{eq1d} limit the sensing and C\&C signal transmit powers within the maximum transmit power $\mathrm{P_{max}}$;
\eqref{eq1e} enforces unit-norm beamforming vectors;
constraint \eqref{eq1f} guarantees the C\&C signal can be successfully decoded at the vehicle; 
constraint \eqref{eq1g} restricts vehicle acceleration within feasible physical limits; 
\eqref{eq1h} guarantees collision avoidance by enforcing that the occupied regions of any two vehicles do not overlap; 
$\mathcal{P}_{1.2}$ minimizes the total number of transmitted sensing and C\&C signals, subject to the same operational and safety constraints.

The optimization problems in $\mathcal{P}_{1.1}$ and $\mathcal{P}_{1.2}$ are challenging to solve directly because of their mixed discrete-continuous decision structure, nonlinear constraints, and long-term temporal coupling across consecutive time slots. As a result, obtaining an exact solution is computationally prohibitive for real-time intersection management. To improve tractability, we decompose the original problems into two hierarchical subproblems.

\noindent\textbf{1) Upper-Level Problem (Signaling and Motion Control):} The upper level governs the macro-management of the intersection. It determines sensing/C\&C transmission scheduling and vehicle acceleration control, with the objectives of maximizing traffic throughput and minimizing the total nunmer of transmitted signals
\begin{align}
    \mathcal{P}_{2.1}:\quad\mathop{\max}_{\substack{\{\delta_n^s(t_i), \delta_n^c(t_i),\\a_{m_k}(t_i)\}_{i=0}^{\mathrm{I}-1}}} &~ \mathbb{E}\Bigg[\sum_{k=1}^{\mathrm{K}} |\mathcal{M}_k(t_{\mathrm I -1})| \Bigg] \label{eq2.1} \\
    \textrm{s. t. } &~\eqref{eq1a},\, \eqref{eq1b},\,\eqref{eq1g},\,\eqref{eq1h}. \notag   \\
    \mathcal{P}_{2.2}:\quad\mathop{\min}_{\substack{\{\delta_n^s(t_i), \delta_n^c(t_i),\\a_{m_k}(t_i)\}_{i=0}^{\mathrm{I}-1}}} &~ \mathbb{E}\Bigg[\sum_{i=0}^{\mathrm{I-1}}  \sum_{n=1}^{\mathrm{N}}(\delta_n^s(t_i)+\delta^c_n(t_i)) \Bigg] \label{eq2.2}  \\
    \textrm{s. t. } &~\eqref{eq1a},\, \eqref{eq1b},\,\eqref{eq1g},\,\eqref{eq1h}. \notag  
\end{align}

\noindent\textbf{2) Lower-Level Problem (Power Allocation and Beamforming):}
Given the signaling decisions $\delta_n^s(t_i)$ and $\delta_n^c(t_i)$ at each time slot, the lower level is activated only when at least one RSU is scheduled to transmit sensing or C\&C signals, i.e., $\sum_{n=1}^{\mathrm{N}}\big(\delta_n^s(t_i)+\delta_n^c(t_i)\big)>0$. 
Its objective is to ensure reliable C\&C signal transmission and obtain high-precision sensing measurements
\begin{align}
        \mathcal{P}_{3}:\quad
        \mathop{\min}_{\substack{p_n^s(t_i), p_n^c(t_i)\\ \mathbf{w}_n^s(t_i), \mathbf{w}_n^c(t_i)}} 
        &~\sum_{k=1}^{\mathrm{K}}\mathrm{Tr}\big(\hat{\mathbf{P}}_{m_k}^G(t_i)\big)  \label{eq3} \\
    \textrm{s. t. } 
        &~\eqref{eq1c}-\eqref{eq1f}. \notag   
\end{align}

\section{Proposed Framework}
\begin{figure*}
    \centering
    \includegraphics[width=0.8\textwidth]{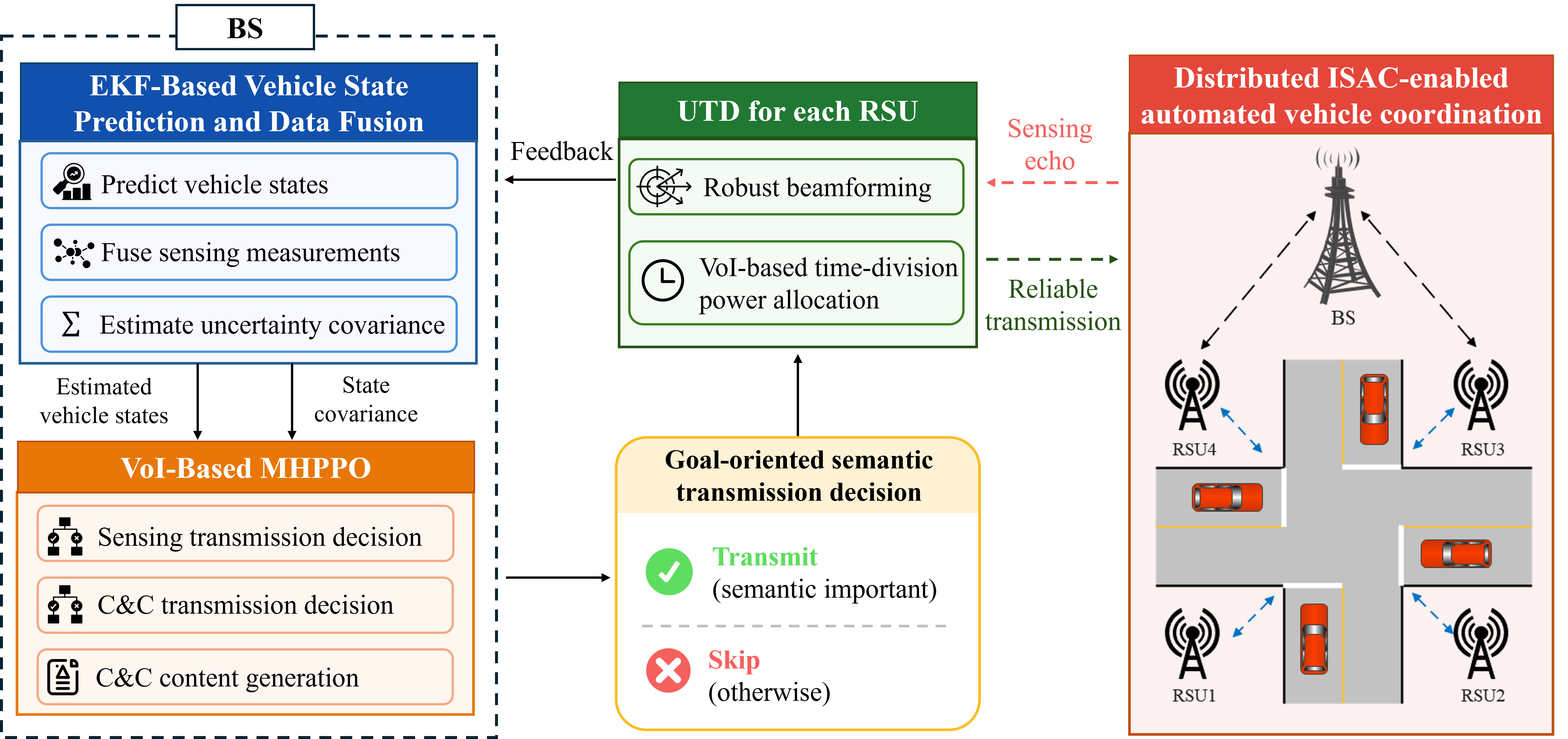}
    \caption{The proposed GSC framework.}
    \label{GSC_framework}
    \vspace{-0.1cm}
\end{figure*}
In this section, we present our proposed unified GSC framework to tackle $\mathcal{P}_{2.1}$, $\mathcal{P}_{2.2}$, and $\mathcal{P}_{3}$. As shown in \figref{GSC_framework}, the framework consists of three tightly coupled components: extended Kalman filter (EKF)-based vehicle state prediction, masked hybrid proximal policy optimization (MHPPO)-based value of information (VoI)-driven decision-making, and uncertainty-aware transmission design (UTD). 
The EKF provides predicted vehicle states and uncertainty covariances at every time slot, which supplies a fundamental basis to reduce unnecessary sensing and C\&C transmissions for $\mathcal{P}_{2.2}$. 
Based on the EKF outputs, MHPPO learns to generate vehicle accelerations and schedule sensing and C\&C transmissions according to a VoI-based reward, such that transmissions are triggered only when they are semantically important for improving traffic throughput, thereby addressing $\mathcal{P}_{2.1}$ and $\mathcal{P}_{2.2}$. 
Given the MHPPO decisions, UTD executes the selected transmissions by accounting for state uncertainty, thereby improving sensing robustness and accuracy while ensuring reliable C\&C communication for $\mathcal{P}_{3}$.
For ease of distinction, the sensing measurement, EKF prediction, and final EKF output are denoted by $\hat{(\cdot)}$, $\breve{(\cdot)}$, and $\mathring{(\cdot)}$, respectively.

\subsection{Extended Kalman Filter for Vehicle State Estimation}
An EKF is deployed at the BS to perform recursive vehicle state estimation. The EKF serves two primary functions: 
(i) predicting vehicle states in the absence of sensing signals, and 
(ii) fusing predicted states with sensing measurements when RSUs provide sensing results. 
To enable recursive estimation, the nonlinear state transition function $f(\mathbf{s}_{m_k}(t_{i-1}),\mathbf{u}_{m_k}(t_{i-1}))$ in \eqref{bicycle_func} is linearized around the previous state estimate, and the corresponding Jacobian matrix $\mathbf{F}_{m_k}(t_{i-1})$ is given by
\begin{equation}
\mathbf{F}_{m_k}(t_{i-1}) =
\begin{bmatrix}
1 & 0 & -\mathring{v}_{i-1}\sin(\mathring{\psi}_{i-1})\Delta t & \cos(\mathring{\psi}_{i-1})\Delta t \\
0 & 1 & \phantom{-}\mathring{v}_{i-1}\cos(\mathring{\psi}_{i-1})\Delta t & \sin(\mathring{\psi}_{i-1})\Delta t \\
0 & 0 & 1 & \dfrac{\tan(\omega_{i-1})}{\mathrm{L_{wb}}}\Delta t \\
0 & 0 & 0 & 1
\end{bmatrix},
\end{equation}
where $\mathring{v}_{i-1} = \mathring{v}_{m_k}(t_{i-1})$, $\mathring{\psi}_{i-1} = \mathring{\psi}_{m_k}(t_{i-1})$, and $\omega_{i-1}=\omega_{m_k}(t_{i-1})$ for notational simplicity. Given the posterior state estimate $\mathring{\mathbf{s}}_{m_k}(t_{i-1})$ and covariance matrix $\mathring{\mathbf{P}}_{m_k}(t_{i-1})$ at time $t_{i-1}$, the a priori prediction at time $t_i$ is computed as
\begin{align}
&\breve{\mathbf{s}}_{m_k}(t_i)= f\big(\mathring{\mathbf{s}}_{m_k}(t_{i-1}), \mathbf{u}_{m_k}(t_{i-1})\big), \\
&\breve{\mathbf{P}}_{m_k}(t_i) = \mathbf{F}_{m_k}(t_{i-1})\,\mathring{\mathbf{P}}_{m_k}(t_{i-1})\,(\mathbf{F}_{m_k}(t_{i-1}))^\top + \mathbf{Q}_{m_k}(t_i),
\end{align}
where $\mathbf{Q}_{m_k}(t_i) \in\mathbb{R}^{4\times4}$ denotes the process noise covariance matrix. After the prediction step, the EKF proceeds differently depending on whether sensing signals are transmitted by RSUs. For notational simplicity, the time index $t_i$ is omitted in the following discussion unless needed to avoid ambiguity.

\noindent \textbf{1) Prediction Case:} If no sensing signals are transmitted at $t_i$, the EKF directly outputs the predicted estimates
\begin{equation}
\mathring{\mathbf{s}}_{m_k} = \breve{\mathbf{s}}_{m_k}, 
\quad
\mathring{\mathbf{P}}_{m_k} = \breve{\mathbf{P}}_{m_k}.
\end{equation}

\noindent \textbf{2) Measurement-Update Case:} When sensing signals are transmitted by RSUs, the BS incorporates the measurement to refine the predicted state. Specifically, the received sensing measurements and their associated uncertainties are first fused at the BS according to~\eqref{mea_fusion_value} and~\eqref{mea_fusion_var}, yielding a global measurement $\hat{\mathbf{s}}^{G}_{m_k}$ and its corresponding covariance matrix $\hat{\mathbf{P}}^{G}_{m_k}$. To ensure consistency between the predicted state and the measurement space, an observation matrix $\mathbf{O} \in \mathbb{R}^{3 \times 4}$ is introduced to extract the observable components, i.e., position and speed, from the full state vector
\begin{equation}
\mathbf{O} =
\begin{bmatrix}
1 & 0 & 0 & 0 \\
0 & 1 & 0 & 0 \\
0 & 0 & 0 & 1
\end{bmatrix}.
\end{equation}

Based on the predicted state and the fused sensing measurement, the EKF performs the update step as follows.

\noindent \textbf{a) Residual:}  
The residual represents the discrepancy between the fused sensing measurement and its predicted value, capturing the new information introduced by the sensing data
\begin{equation}
\mathbf{r}_{m_k} = \hat{\mathbf{s}}^{G}_{m_k} - \mathbf{O}\,\breve{\mathbf{s}}_{m_k}.
\end{equation}

\noindent \textbf{b) Residual Covariance:}  
The residual covariance quantifies the uncertainty of $\mathbf{r}_{m_k}$ by combining the predicted state uncertainty and the sensing measurement uncertainty as follows
\begin{equation}
\mathbf{S}_{m_k} = \mathbf{O}\,\breve{\mathbf{P}}_{m_k}\,\mathbf{O}^\top + \hat{\mathbf{P}}^{G}_{m_k}.
\end{equation}

\noindent \textbf{c) Kalman Gain:}  
The Kalman gain determines the relative weighting between the predicted state and the sensing measurement
\begin{equation}
\mathbf{G}_{m_k} = \breve{\mathbf{P}}_{m_k}\,\mathbf{O}^\top\,(\mathbf{S}_{m_k})^{-1}.
\end{equation}

\noindent \textbf{d) State and Covariance Update:}  
The final state estimate is obtained by correcting the predicted state using the residual weighted by the Kalman gain
\begin{equation}
\mathring{\mathbf{s}}_{m_k} = \breve{\mathbf{s}}_{m_k} + \mathbf{G}_{m_k}\,\mathbf{r}_{m_k}.
\end{equation}

The covariance matrix of $\mathring{\mathbf{s}}_{m_k}$ is then updated as
\begin{equation}
\mathring{\mathbf{P}}_{m_k} = \big(\mathbf{I}_4 - \mathbf{G}_{m_k}\mathbf{O}\big)\,\breve{\mathbf{P}}_{m_k}.
\end{equation}



\subsection{Masked Hybrid Proximal Policy Optimization for Sensing and C\&C Signal Transmission}
Based on the EKF-estimated vehicle states and covariances, we develop an MHPPO framework to jointly determine sensing and C\&C signal transmissions as well as vehicle acceleration. Unlike standard PPO, MHPPO manages a hybrid action space by simultaneously parameterizing categorical distributions for discrete transmission decisions and Gaussian distributions for continuous acceleration. To enforce the inherent causality between sensing and C\&C signaling  (i.e., $\delta_n^c(t_{i+1}) \leq \delta_n^s(t_{i})$), we integrate a masking mechanism to filter out infeasible actions, ensuring that invalid C\&C decisions and their associated acceleration outputs are not selected. Consistent with the upper-level problem, the state, action, and reward of MHPPO are defined as follows.

\noindent \textbf{1) State:} The observation state at time slot $t_i$ is defined as
\begin{equation}
\mathcal{S}(t_i)\triangleq \Big\{\mathbf d^{\text{ext}},\,\mathbf D^{\text{coll}},\, \mathring{\mathbf v}, \,\mathbf{p}^{\mathrm{tr}},\,N_{\delta}(t_i),\,N_{\text{veh}}(t_i)\Big\},
\end{equation}
where $\mathbf{d}^{\text{ext}} = [d^{\text{ext}}_{m_1}, \ldots, d^{\text{ext}}_{m_{\mathrm{K}}}]^\top$ denotes the distances from vehicles to their exit points, calculated based on the EKF-estimated vehicle positions; $\mathbf{D}^{\text{coll}} = [(\mathbf{d}^{\text{coll}}_{m_1})^\top, \ldots, (\mathbf{d}^{\text{coll}}_{m_{\mathrm{K}}})^\top]^\top$ captures the distances to potential collision areas pre-computed from fixed vehicle routes, where each vector $\mathbf{d}^{\text{coll}}_{m_k}$ is determined by the EKF-estimated position relative to the corresponding collision areas along the route of vehicle $m_k$; $\mathring{\mathbf{v}} = [\mathring{v}_{m_1}, \ldots, \mathring{v}_{m_{\mathrm{K}}}]^\top$ represents the estimated vehicle speeds; $\mathbf{p}^{\mathrm{tr}}=[\mathrm{Tr}(\mathring{\mathbf{P}}_{m_1}),\ldots, \mathrm{Tr}(\mathring{\mathbf{P}}_{m_{\mathrm{K}}})]^\top$ quantifies vehicle state uncertainties; $N_{\delta}(t_i)= \sum_{j=0}^{i}\sum_{n=1}^{\mathrm{N}}\big(\delta_n^s(t_j)+\delta_n^c(t_j)\big)$ is the cumulative number of transmitted sensing and C\&C signals up to $t_i$; $N_{\text{veh}}(t_i)=\sum_{k=1}^{\mathrm{K}}|\mathcal{M}_k(t_i)|$ is the total number of vehicles successfully passed the intersection up to $t_i$.


\noindent \textbf{2) Action:} The action space at time slot $t_i$ is composed of the decisions of all RSUs, i.e., $\mathcal{A}(t_i) = \{\mathcal{A}_n(t_i):n=1,\dots, \mathrm{N}\}$. Since each RSU shares an identical action structure, the per-RSU action space is defined as
\begin{equation}
\mathcal{A}_n(t_i) \triangleq \{A_n^d,~a_{m_k}\},
\end{equation}
where $A_n^d \in \{0,1,2\}$ denotes the discrete action that jointly determines $\delta_n^s$ and $\delta_n^c$: $A_n^d=0$ corresponds to silence, $A_n^d=1$ indicates sensing transmission only, and $A_n^d=2$ represents transmitting the sensing signal at $t_i$ and the C\&C signal at $t_{i+1}$. The continuous action $a_{m_k}$ specifies the vehicle acceleration, constrained within the interval $[\mathrm{A_{min}}, \mathrm{A_{max}}]$. A masking mechanism is introduced such that $a_{m_k}(t_{i+1})$ is activated only when $A_n^d(t_{i}) = 2$. This action space design ensures that the constraints in \eqref{eq1a}, \eqref{eq1b}, and \eqref{eq1g} are satisfied.

\noindent \textbf{3) Reward:} The reward function at time slot $t_i$ is defined as
\begin{equation}
R(t_i)\triangleq \text{VoI} - \text{Cost} + \Psi_{\text{pass}} - \Psi_{\text{coll}},
\end{equation}
where each term is specified as follows.

\noindent\textbf{a) VoI:} The value of information (VoI) encourages the BS to learn to transmit sensing and C\&C signals when they are critical to the task, which consists of sensing and C\&C  contributions
\begin{equation}
\text{VoI} \triangleq \text{VoI}_s + \text{VoI}_c,
\end{equation}
in which the sensing VoI is quantified by the reduction in estimation uncertainty
\begin{equation}
\text{VoI}_s \triangleq \sum_{k=1}^{\mathrm{K}} \mathrm{Tr}\big(\mathring{\mathbf{P}}_{m_k}(t_{i-1})\big) - \mathrm{Tr}\big(\mathring{\mathbf{P}}_{m_k}(t_i)\big),
\end{equation}
and the C\&C VoI is defined by the reduction in remaining distance to the exit point
\begin{equation}
\text{VoI}_c \triangleq d_{m_k}^{\text{ext}}(t_{i+\mathrm{I_w}\Delta t} \,|\, \delta_n^c=0) - d_{m_k}^{\text{ext}}(t_{i+\mathrm{I_w}\Delta t} \,|\, \delta_n^c=1),
\end{equation}
where $\mathrm{I_w}\Delta t$ is a look-ahead time window capturing the long-term impact of acceleration.

\noindent\textbf{b) Cost:} To incorporate the objective to reduce the number of transmitted sensing and C\&C signals in \eqref{eq2.2}, the transmission cost for sensing and C\&C signals is defined as
\begin{equation}
\text{Cost} \triangleq
\begin{cases}
0.5, & \text{if } A_n^d=1, \\
0.5\cdot(1+\mathrm{N_c/N_s}), & \text{if } A_n^d=2, \\
0, & \text{otherwise}.
\end{cases}
\end{equation}

\noindent\textbf{c) Passing Reward:} To encourage higher traffic throughput as in \eqref{eq2.1}, the reward for successfully passing vehicles is 
\begin{equation}
\Psi_{\text{pass}} \triangleq \Upsilon_{\text{pass}} \big(|\mathcal{M}_k(t_i)| - |\mathcal{M}_k(t_{i-1})|\big).
\end{equation}

\noindent\textbf{d) Collision Penalty:} A large penalty is imposed for violating the safety constraint in \eqref{eq1h}
\begin{equation}
\Psi_{\text{coll}}\triangleq
\begin{cases}
\Upsilon_{\text{coll}}, & \text{if } \mathcal{C}(\mathbf{s}_{m_k}) \cap \mathcal{C}(\mathbf{s}_{m_{\widetilde{k}}}) \neq \varnothing, \\
0, & \text{otherwise}.
\end{cases}
\end{equation}

With the above definitions, the upper-level problem can be formulated as the following Markov decision process (MDP)
\begin{equation}
\mathcal{P}_4:\quad \max_{\pi_{\bm\theta}(\mathcal{A}(t_i)|\mathcal{S}(t_i))} \mathbb{E}_{\pi_{\bm\theta}} \Bigg[ \sum_{i=0}^{\mathrm{I-1}} \gamma^i R(t_i) \Bigg],
\end{equation}
where $\pi_{\bm\theta}$ denotes the stochastic policy network parameterized by $\bm\theta$, $\gamma\in(0,1)$ is the discount factor. 
We establish an MHPPO framework to solve the MDP. Specifically, MHPPO adopts an actor--critic architecture, where the actor outputs a categorical distribution for the discrete action and a Gaussian distribution for the continuous action. The continuous action is regulated to $[\mathrm{A_{min}}, \mathrm{A_{max}}]$ via a `$\tanh$' squashing function followed by affine scaling. The mask mechanism is introduced such that the log-probability contribution of the continuous action is only activated when the selected discrete action $A^d_n = 2$. Otherwise, the continuous branch is ignored and its contribution to the policy gradient is masked out. The masked log-probability for RSU $n$ is given as
\begin{equation}
\begin{aligned}
    \log \pi_{\bm{\theta}}(\mathcal A_n | \mathcal{S}) 
    = & \log P(A^d_n | \mathcal{S})  + \mathds{1}_{A^d_n=2} \bigg[
    \log \mathcal{N}(a_{m_k}^{\text{raw}}| \mu_{\bm{\theta}}, \sigma_{\bm{\theta}})  \\
    & - \log\Big(1 - \Big(\tanh(a_{m_k}^{\text{raw}})\Big)^2\Big)
    \bigg],
\end{aligned}
\end{equation}
where the first term on the RHS denotes the log-probability of the discrete action under a categorical distribution, the second term denotes the masked log-probability of the continuous action under a Gaussian distribution, $\mathds{1}_{A^d_n=2}$ is an indicator function that equals $1$ if $A^d_n = 2$ and $0$ otherwise, $a_{m_k}^{\text{raw}} \in (-\infty, +\infty)$ denotes the raw continuous action sampled from the Gaussian distribution with mean $\mu_{\bm{\theta}}$ and standard deviation $\sigma_{\bm{\theta}}$. The critic outputs the state value, which represents the expected cumulative discounted reward under the current policy. Specifically, the value function is defined as
\begin{equation}
    V_{\pi_{\bm\theta}}(\mathcal{S}(t_i)) 
    = \mathbb{E}_{\pi_{\bm\theta}}\left[
    \sum_{\ell=0}^{\mathrm{I-1}-\ell} \gamma^\ell R(t_{i+\ell}) 
    \;\middle|\; \mathcal{S}(t_i)
    \right].
\end{equation}

The exploration and training process of the MHPPO framework is described as follows. At time step $t_i$, the agent observes the environment state $\mathcal{S}(t_i)$ and feeds it into the actor--critic network to obtain the action $\mathcal{A}(t_i)$ and the state value $V_{\pi_{\bm\theta}}(\mathcal{S}(t_i))$. The agent then executes $\mathcal{A}(t_i)$, receives reward $R(t_i)$, and the environment transitions to the next state $\mathcal{S}(t_{i+1})$. The transition tuple
$\big(\mathcal{S}(t_i), \mathcal{A}(t_i), R(t_i), \log \pi_{\bm{\theta}}(\mathcal{A}(t_i) | \mathcal{S}(t_i)),\mathcal{S}(t_{i+1})\big)$ is stored in a replay buffer.
After one episode of exploration, the MHPPO framework updates the actor and critic networks using the stored experiences. Let the collected trajectory be denoted as 
\begin{equation}
    \big\{(\mathcal{S}(t_i), \mathcal{A}(t_i), R(t_i), \log \pi_{\bm{\theta}}(\mathcal{A}(t_i)|\mathcal{S}(t_i)), V_{\pi_{\bm{\theta}}}(\mathcal{S}(t_i)))\big\}_{i=0}^{\mathrm{I}-1}.
\end{equation}

To reduce variance while maintaining low bias, the generalized advantage estimation (GAE) is adopted. The temporal-difference (TD) error for trajectory at $t_i$ is defined as
\begin{equation}
    \varpi_i = R(t_i) + \gamma V_{\pi_{\bm{\theta}}}(\mathcal{S}(t_{i+1})) - V_{\pi_{\bm{\theta}}}(\mathcal{S}(t_i)).
\end{equation}

The advantage function is computed recursively as
\begin{equation}
    \Gamma_i = \varpi_i + \gamma \lambda \Gamma_{i+1},
\end{equation}
where $\Gamma_i$ is the advantage value initialized as $\Gamma_0=0$, and $\lambda\in(0,1)$ is the GAE parameter. The estimated return is then given by
\begin{equation}
    \hat{R}_i = \Gamma_i + V_{\pi_{\bm{\theta}}}(\mathcal{S}(t_i)).
\end{equation}

Based on the advantage values $\{\Gamma_i\}_{i=0}^{\mathrm{I-1}}$, the actor network is updated by maximizing the clipped surrogate objective. The probability ratio is defined as
\begin{equation}
    r_i(\bm{\theta}) = \frac{\pi_{\bm{\theta}}(\mathcal{A}(t_i)|\mathcal{S}(t_i))}{\pi_{\bm{\theta}_{\text{old}}}(\mathcal{A}(t_i)|\mathcal{S}(t_i))}
    = \exp\Big( \log \pi_{\bm{\theta}} - \log \pi_{\bm{\theta}_{\text{old}}} \Big).
\end{equation}
where $\bm{\theta}_{\text{old}}$ denotes the parameters of the actor network before
the current update. The clipped surrogate objective for the actor network is given by
\begin{equation}
\begin{aligned}
    \mathcal{L}_{\text{act}}(\bm{\theta}) = 
    \mathbb{E}_i \Big[
    \min \big( 
    r_i(\bm{\theta}) \Gamma_i, \;
    \text{clip}(r_i(\bm{\theta}), 1-\epsilon, 1+\epsilon)\Gamma_i
    \big)
    \Big],
\end{aligned}
\end{equation}
where $\epsilon\in(0,1)$ is the clipping parameter. Based on the estimated returns $\{\hat{R}_i\}_{i=0}^{\mathrm{I-1}}$, the critic network is trained by minimizing the mean squared error between the predicted state value and the estimated return
\begin{equation}
    \mathcal{L}_{\text{crit}}(\bm{\theta}) = 
    \mathbb{E}_i \Big[
    \big( V_{\pi_{\bm{\theta}}}(\mathcal{S}(t_i)) - \hat{R}_i \big)^2
    \Big].
\end{equation}

To encourage exploration, an entropy bonus is incorporated. Considering the masked hybrid action space, the entropy term is defined as
\begin{equation}
\begin{aligned}
    \mathcal{H}_i = \sum_{n=1}^{\mathrm{N}} \Big[
    \mathcal{H}\big(P(A^d_n|\mathcal{S})\big) 
    + \mathds{1}_{A^d_n=2} \mathcal{H}\big(\mathcal{N}(a_{m_k}^{\text{raw}}|\mu_{\bm{\theta}}, \sigma_{\bm{\theta}})\big)
    \Big],
\end{aligned}
\end{equation}
where $\mathcal{H}(\cdot)$ denotes Shannon entropy of the corresponding distribution. The final optimization objective of MHPPO is given by
\begin{equation}
\begin{aligned}
    \mathcal{L}(\bm{\theta}) = 
    - \mathcal{L}_{\text{act}}(\bm{\theta})
    + c_1 \mathcal{L}_{\text{crit}}(\bm{\theta})
    - c_2 \mathbb{E}_i[\mathcal{H}_i],
\end{aligned}
\end{equation}
where $c_1$ and $c_2$ are weighting coefficients for the value loss and entropy bonus, respectively.
The parameters $\bm{\theta}$ are updated via stochastic gradient descent over multiple epochs using mini-batches sampled from the replay buffer with optional gradient clipping for training stability. 

\subsection{Uncertainty-Aware Transmission Design}
The UTD consists of two key components: robust beamforming and VoI-based time-division power allocation. The former improves transmission reliability under vehicle-state uncertainty, while the latter employs a VoI-based time-division mechanism to allocate transmit power and mitigate inter-RSU interference.
\subsubsection{Robust Beamforming}
We adopt the beampattern synthesis approach proposed in \cite{BF} to design the robust transmit beamforming vector $\mathbf{w}_n$, where $\mathbf{w}_n$ comprises both the sensing beamforming vector $\mathbf{w}_n^s$ and the communication beamforming vector $\mathbf{w}_n^c$. Specifically, a uniform angular grid spanning the interval $\left[-\frac{\pi}{2}, \frac{\pi}{2}\right]$ is constructed with $\mathrm{N}_{\theta}$ discrete angular samples, denoted by $\{\theta_u\}_{u=1}^{\mathrm{N}_{\theta}}$. 
Based on the estimated AoA $\hat{\theta}_n^{m_k}$ and its associated estimation variance $\sigma^2(\hat{\theta}_n^{m_k})$, the angular scan scope from the $n$-th RSU toward vehicle $m_k$ is defined as $\bm{\theta}_{\mathrm{SS}}^n=[\theta_{\min}^n,\theta_{\max}^n],$
where the boundary angles are determined by
\begin{equation}
    \begin{cases}
        \theta_{\min}^n = \hat{\theta}_n^{m_k} - \mathrm{B}\sigma(\hat{\theta}_n^{m_k}), \\ 
        \theta_{\max}^n = \hat{\theta}_n^{m_k} + \mathrm{B}\sigma(\hat{\theta}_n^{m_k}),
    \end{cases}
\end{equation}
with $\mathrm{B}$ denoting the confidence-level scaling factor. A larger value of $\mathrm{B}$ yields a wider angular coverage, thereby improving robustness against AoA estimation uncertainty. Let $\mathbf{b}_n \in \mathbb{C}^{\mathrm{N}_{\theta}\times 1}$ denote the desired beam pattern over the angular grid. Its $u$-th entry is defined as
\begin{equation}
[\mathbf{b}_n]_u
\triangleq
\begin{cases}
\mathrm{K},
& \theta_u \in \bm{\theta}_{\mathrm{SS}}^n,
\\[1mm]
0,
& \text{otherwise},
\end{cases}
\label{eq:desired_pattern}
\end{equation}

The robust beamforming design is formulated as the following least-squares beampattern matching problem
\begin{equation}
\min_{\mathbf{w}_n}\left\|\mathbf{b}_n-\mathbf{A}_{\mathrm{t}}^\top \mathbf{w}_n\right\|_2^2,
\end{equation}
where $\mathbf{A}_{\mathrm{t}}=[\mathbf{a}_{\mathrm{t}}(\theta_1),\ldots,\mathbf{a}_{\mathrm{t}}(\theta_{\mathrm{N}_{\theta}})]$ is the transmit steering matrix evaluated over the predefined angular grid, and $\mathbf{a}_{\mathrm{t}}(\theta_u)$ denotes the transmit steering vector corresponding to angle $\theta_u$. The least-squares problem admits the closed-form solution
\begin{equation}
\mathbf{w}_n=(\mathbf{A}_{\mathrm{t}}^*\mathbf{A}_{\mathrm{t}}^\top)^{-1}\mathbf{A}_{\mathrm{t}}^*\mathbf{b}_n.
\end{equation}

The normalized beamforming vector is obtained as follows
\begin{equation}
\mathbf{w}_n
\leftarrow
\frac{\mathbf{w}_n}{\|\mathbf{w}_n\|_2}.
\label{eq:bf_normalization}
\end{equation}

\subsubsection{VoI-Based Time-Division Power Allocation}
For sensing transmission, all RSUs can simultaneously transmit sensing signals, as the orthogonal sensing waveforms effectively avoid inter-cell interference. In contrast, concurrent C\&C transmissions from multiple RSUs may cause severe inter-cell interference. To mitigate this issue, we propose a VoI-based time-division power allocation strategy, where C\&C transmissions with higher VoI are assigned higher scheduling priority.
The time-division design is motivated by the following practical observations. The time-slot duration is $\Delta t=5$ ms, while the OFDM symbol duration under 60 kHz subcarrier spacing is approximately $\mathrm{T_{sym}} = 18~\upmu$s. Due to the small payload of the C\&C data, it may only occupy a few OFDM symbols (i.e.,  $\mathrm{N}^c_{\mathrm{sym}}\leq7$),  leaving sufficient temporal resources within one time slot for fine-grained time-division scheduling. Moreover, 5G New Radio supports flexible half-slot (i.e., 7 OFDM symbols) scheduling, which allows different RSUs to transmit over different half-slots within the same time slot.

According to the actions generated by the MHPPO policy, each RSU operates in one of three modes: sensing-only, communication-only, or ISAC transmission. The corresponding VoI-based time-division power allocation strategies are detailed below.

\noindent \textbf{a) Sensing-Only Transmission:} In this mode, the $n$-th RSU allocates its entire power budget to sensing for the duration of the slot, such that
\begin{equation}
    p_n^{s}(t_i) = \mathrm{P_{max}} , \quad p_n^{c}(t_i) = 0.
\end{equation}

\noindent \textbf{b) Communication-Only Transmission:} The scheduling for communication depends on the activity of neighboring RSUs to mitigate interference:
\begin{itemize}
    \item \textbf{Case 1: Single-RSU Transmission.} If no other RSUs are performing communication, the $n$-th RSU occupies the first half-slot
    \begin{equation}
        p_n^{c}(t_i) = 
        \begin{cases} 
            \mathrm{P_{max}}, & i\Delta t \le t_i \le i\Delta t + 6\mathrm{T}_{\mathrm{sym}}, \\
            0, & \text{otherwise}.
        \end{cases}
    \end{equation}
    \item \textbf{Case 2: Multi-RSU Scheduling.} If multiple RSUs require communication, links are prioritized by their VoI. Higher-VoI links are assigned earlier OFDM symbol intervals. For example,  if the $n$-th RSU is ranked second in VoI, its transmit power is given by
    \begin{equation}
    \begin{aligned}
        &p_n^{c}(t_i)= \\ 
        &\begin{cases} 
            \mathrm{P_{max}}, & i\Delta t + 14\mathrm{T_{sym}} \le t_i \le i\Delta t + 20\mathrm{T_{sym}}, \\ 
            0, & \text{otherwise}.
        \end{cases}
    \end{aligned}
    \end{equation}
\end{itemize}

\noindent \textbf{c) ISAC Transmission:} In the ISAC mode, the $n$-th RSU aims to maximize sensing performance while satisfying communication requirements. Since the time-division scheduling eliminates inter-RSU interference, the SINR constraint in \eqref{eq1f} reduces to a signal-to-noise ratio (SNR) requirement
\begin{equation}
    \frac{\mathrm{N_t} p_n^{c}(t_i)|\kappa_n^{m_k}|^2\left|\mathbf{a}_{\mathrm{t}}^{\mathrm{H}}(\theta_n^{m_k})
    \mathbf{w}_n^{c}\right|^2}{\sigma^2_{\text{clt}} + \sigma_c^2} \geq \gamma_{\mathrm{thr}}.
\end{equation}

Accordingly, the minimum communication transmit power is obtained as
\begin{equation} \label{p_nc(t)}
    p_n^{c}(t_i)=\frac{\gamma_{\mathrm{thr}}(\sigma^2_{\text{clt}} + \sigma_c^2)}{\mathrm{N_t}|\kappa_n^{m_k}|^2\left|\mathbf{a}_{\mathrm{t}}^{\mathrm{H}}(\theta_n^{m_k})
    \mathbf{w}_n^{c}\right|^2}.
\end{equation}

The residual power, $p_n^{s}(t_i) = \mathrm{P_{max}} - p_n^{c}(t_i)$, is allocated to sensing. Scheduling follows a similar time-division strategy depends on the activity of neighboring RSUs:
\begin{itemize}
    \item \textbf{Case 1: Single-RSU Transmission.} Communication occurs during the first half-slot ($i\Delta t \leq t_i \leq i\Delta t + 6\mathrm{T_{sym}}$), while sensing is performed across the entire slot. During this specific interval, communication transmit power is calculated by \eqref{p_nc(t)}, while sensing uses the remaining power $\mathrm{P_{max}} - p_n^c(t_i)$; during all other intervals, sensing utilizes the full power $\mathrm{P_{max}}$.
    \item \textbf{Case 2: Multi-RSU Scheduling.} Communication is scheduled in a specific interval based on VoI ranking (e.g., from the 14th to the 20th symbol). During this specific interval, communication transmit power is calculated by \eqref{p_nc(t)}, while sensing uses the remaining power $\mathrm{P_{max}} - p_n^c(t_i)$; during all other intervals, sensing utilizes the full power $\mathrm{P_{max}}$.
\end{itemize}

\section{Simulation and Analysis}
In this section, extensive simulations are conducted to evaluate the performance of our proposed GSC framework. The key simulation parameters are summarized as follows.
The time slot interval is set to $\Delta t = 5$ ms, and the total number of time slots is  $\mathrm{I}=12000$. We leverage Simulation of Urban MObility (SUMO) platform \cite{SUMO} to construct the intersection scenario, generate vehicle traffic, and control vehicle motion. Specifically,
the intersection consists of $\mathrm{K}=4$ roads, where the control area of each road has a side length of $\mathrm{L_c}=4.8$ m. The conflict area is modeled as a square region with a side length of $14.4$ m. The maximum vehicle speed when passing through the intersection is set to $8.3$ m/s and the maximum accelerations is $\mathrm{A_{max}}=5$ m/s$^2$. The length and width of each vehicle are $\mathrm{L}=4.6$ m and $\mathrm{W}=1.8$ m, respectively. 
For the wireless simulation setup, a total of $\mathrm{N}=4$ RSUs are deployed around the intersection, with locations $[-15, -20]^\top\,\text{m}$, $[20, -15]^\top\,\text{m}$, $[15, 20]^\top\,\text{m}$, and $[-20, 15]^\top\,\text{m}$, respectively.
Each RSU is equipped with $\mathrm{N_t}=32$ transmit antennas and $\mathrm{N_r}=32$ receive antennas, and the maximum transmit power is $\mathrm{P_{max}}=200$ mW. Each RSU uses $\mathrm{N_s}=2500$ subcarriers for sensing and $\mathrm{N_c}=50$ subcarriers for C\&C signal transmission, with a subcarrier spacing of $\Delta f =60$~kHz. The carrier frequency is $f_c = 60$~GHz, and the antenna element spacing is $\mathrm{D} = 2.5$~mm. Within each time slot, the numbers of OFDM symbols allocated for sensing and communication are $\mathrm{N}^s_{\mathrm{sym}}=98$ and $\mathrm{N}^c_{\mathrm{sym}}=3$, respectively, and the duration of one OFDM symbol is $\mathrm{T_{sym}}=18~\upmu$s.
The RCS of each vehicle is set to $\varepsilon_n^{m_k} = 20$ m$^2$. The number of stationary scatterers for sensing is randomly selected from $5 \leq Q \leq 20$, and the average power of the $q$-th scatterer follows $\sigma^2_q\in[-87, -77]$ dB. The power spectrum density of AWGN is $\sigma^2_0=-174$ dBm. The detection parameters are $\alpha_1=0.01$, $\alpha_2=6.7\times10^{-9}$, and $\alpha_3=200$. 
For C\&C signal transmission, the clutter noise power is $\sigma^2_{\mathrm{clt}}\in [-106,-101]$ dB, and the communication threshold is $\gamma_{\text{thr}}=8$ dB.
The look-ahead time slot for $\text{VoI}_c$ is set to $\mathrm{I_w}=20$. 
For robust beamforming, the confidence-level scaling factor is set to $\mathrm{B}=2.576$, corresponding to a $99\%$ confidence level.
All simulations are conducted over $50$ random seeds, and the results are averaged over these independent runs. The hyperparameters of MHPPO are summarized in \textbf{\tabref{tab}}.

\begin{table}[h]
    \renewcommand\arraystretch{1.3}
    \caption{MHPPO Hyperparameter Settings}\label{tab}
    \centering
    \begin{tabular}{|p{5cm}<{\centering}|p{2cm}<{\centering}|}
        \hline
        \textbf{Parameter description}       & \textbf{Value}     \\ 
        \hline
        Collision penalty $\Upsilon_{\text{coll}}$      & 50   \\
        \hline
        Passing reward $\Upsilon_{\text{pass}}$     	 & 10   \\
        \hline
        Discount factor $\gamma$             & 0.99  \\
        \hline
        GAE parameter $\lambda$             & 0.95 \\
        \hline
        Clipping parameter $\epsilon$   & 0.2 \\
        \hline
        Value loss factor $c_1$  &  0.5  \\
        \hline
        Entropy bonus factor $c_2$   & 0.01 \\ 
        \hline
    \end{tabular}
\end{table}

Since this is the first work to investigate distributed ISAC-enabled vehicle coordination at intersections, existing methods cannot be directly applied to the considered scenario. The most related studies are the predictive ISAC transmission schemes in \cite{Liu, Du, extended_target}, which focus on vehicle tracking and downlink communication with a single transmitter and do not involve vehicle control. For fair comparison, we adapt their core design principles to our distributed ISAC-enabled intersection scenario. In addition, we construct several ablation baselines to evaluate the contribution of each key component in our proposed unified GSC framework. The considered comparison algorithms are summarized as follows.

\begin{itemize}
    \item \textbf{Predictive ISAC with PPO (P-ISAC):} 
    This baseline is adapted from the predictive ISAC transmission schemes in \cite{Liu, Du, extended_target}. The EKF is used to predict vehicle states, and both sensing and C\&C signals are transmitted at every time slot. ISAC beamforming is designed based on the EKF-predicted vehicle states, and power allocation aims to improve sensing performance while satisfying communication threshold. Since the original predictive ISAC schemes do not generate vehicle control commands, PPO is used to determine vehicle acceleration with the same hyperparameter settings as the proposed MHPPO.

    \item \textbf{GSC without MHPPO (w/o MHPPO):} 
    This scheme removes the MHPPO module from our proposed GSC framework. Sensing and C\&C signals are transmitted periodically instead of being scheduled according to their VoI-driven semantic importance.
    
    \item \textbf{GSC without UTD (w/o UTD):} 
    This scheme removes the UTD module from our proposed GSC framework. Beamforming is designed only according to the estimated vehicle positions without covariance information, and C\&C signals from different RSUs are transmitted simultaneously.

    \item \textbf{GSC without EKF (w/o EKF):} 
    This scheme removes the EKF module from our proposed GSC framework. When sensing signals are not transmitted, vehicle states are not predicted. Instead, MHPPO observes the most recently sensed vehicle states and the elapsed time slots since the latest sensing transmission.
\end{itemize}

\begin{figure}
    \centering
    \includegraphics[width=0.35\textwidth]{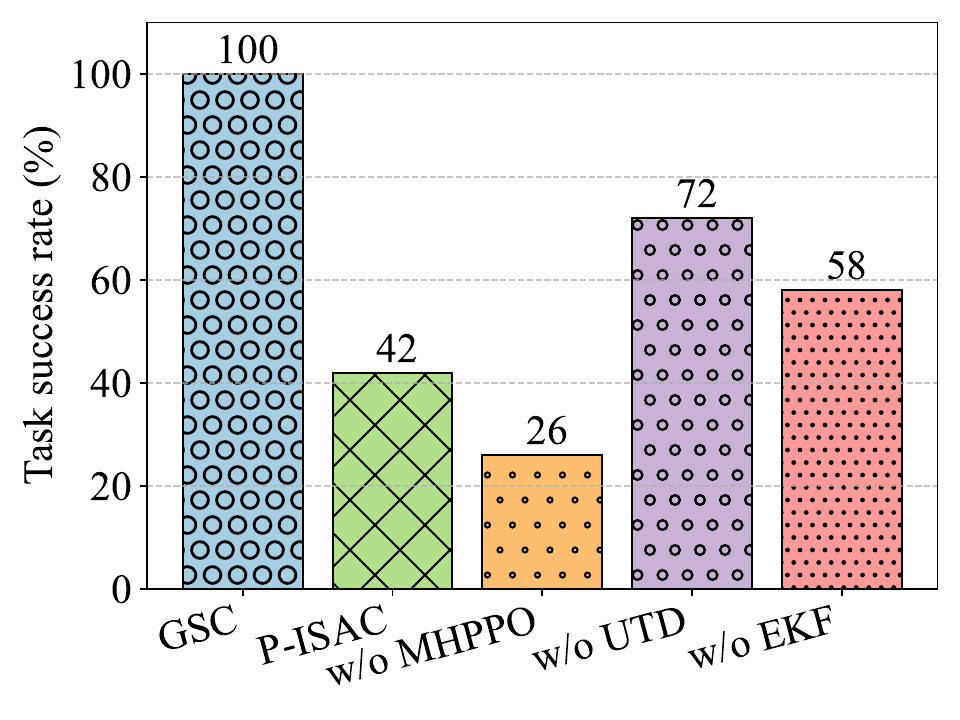}
    \caption{Task success rate.}
    \label{success_rate}
    \vspace{-0.3cm}
\end{figure}

\figref{success_rate} plots the task success rate of different schemes, where task success means that all vehicles pass through the intersection without collision during the entire I time slots. As can be seen, our proposed GSC framework achieves a 100\% task success rate, significantly outperforming all baselines and verifying the effectiveness of our integrated architecture. Although the adapted P-ISAC scheme transmits both sensing and C\&C signals at every time slot, it achieves only a 42\% success rate. This is mainly because its single-transmitter design causes severe inter-RSU interference when extended to the distributed RSU scenario, making C\&C signals difficult to decode reliably. The ablation results in \figref{success_rate} further verify the contribution of each key module. Removing MHPPO leads to the lowest success rate of 26\%, demonstrating the importance of VoI-driven transmission scheduling. Without UTD, the success rate decreases to 72\%. Although this scheme still benefits from EKF-based state prediction and MHPPO-based scheduling, the lack of uncertainty-aware and interference-aware transmission design limits its safety performance. Without EKF, the success rate drops to 58\%, since the system relies on stale vehicle states when sensing signals are not transmitted. 

\begin{figure}[t] 
    \centering
    \subfloat[][Total number of passed vehicles.]
    {\includegraphics[width=0.23\textwidth]{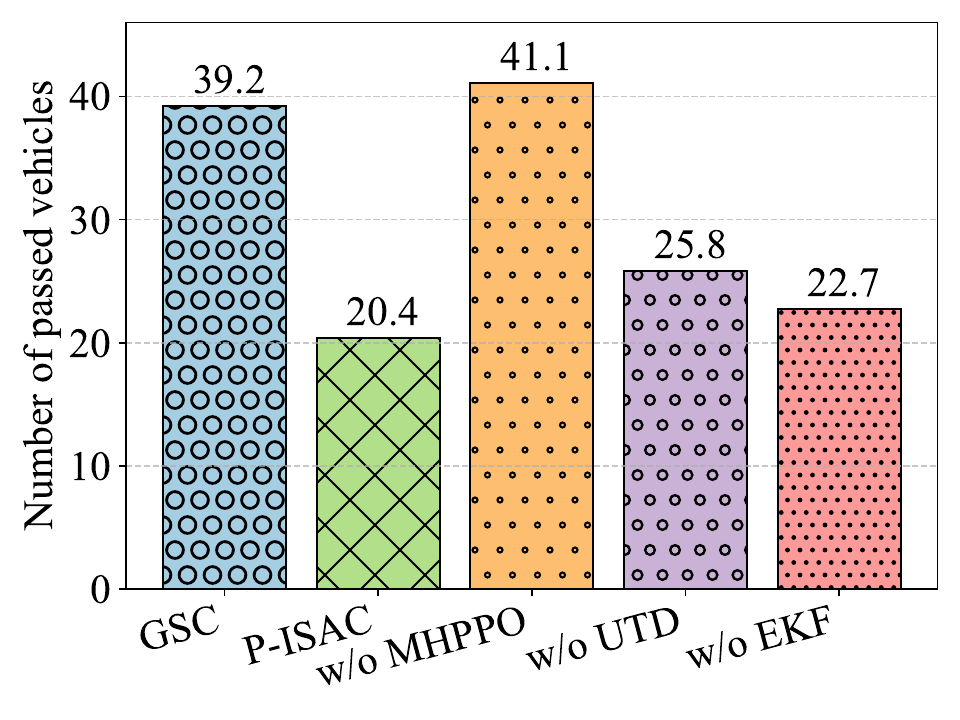}
        \label{total_vehicles}
    }  
    \subfloat[][Total number of transmitted signals for each vehicle.]
    {\includegraphics[width=0.23\textwidth]{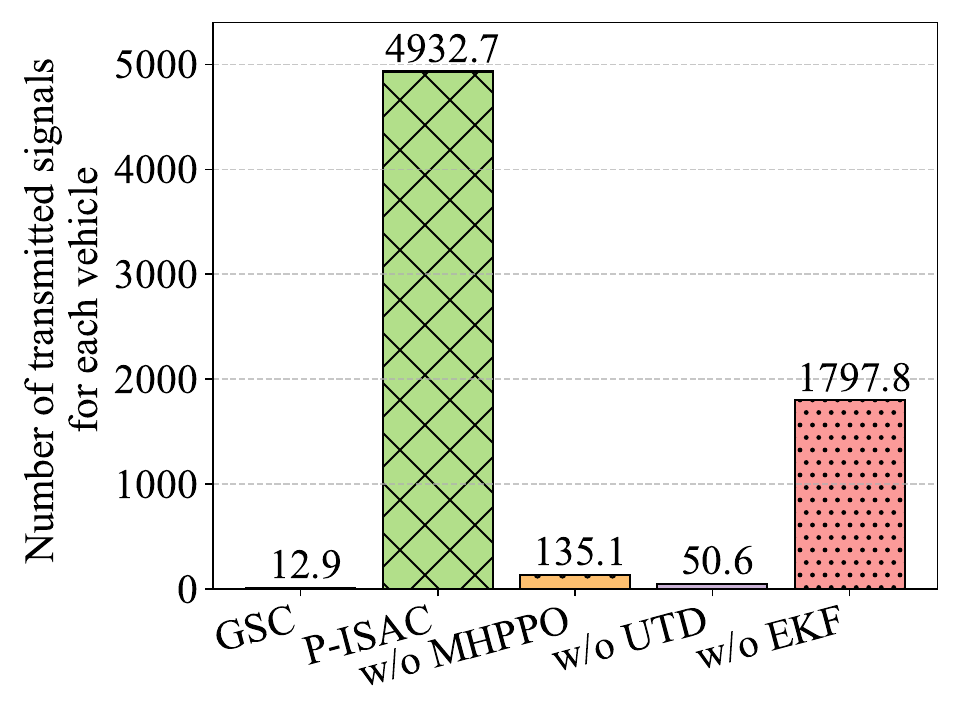}
        \label{signals_per_vehicle}
    }  \\
    \subfloat[][Total number of transmission slots for each RSU.]
    {\includegraphics[width=0.23\textwidth]{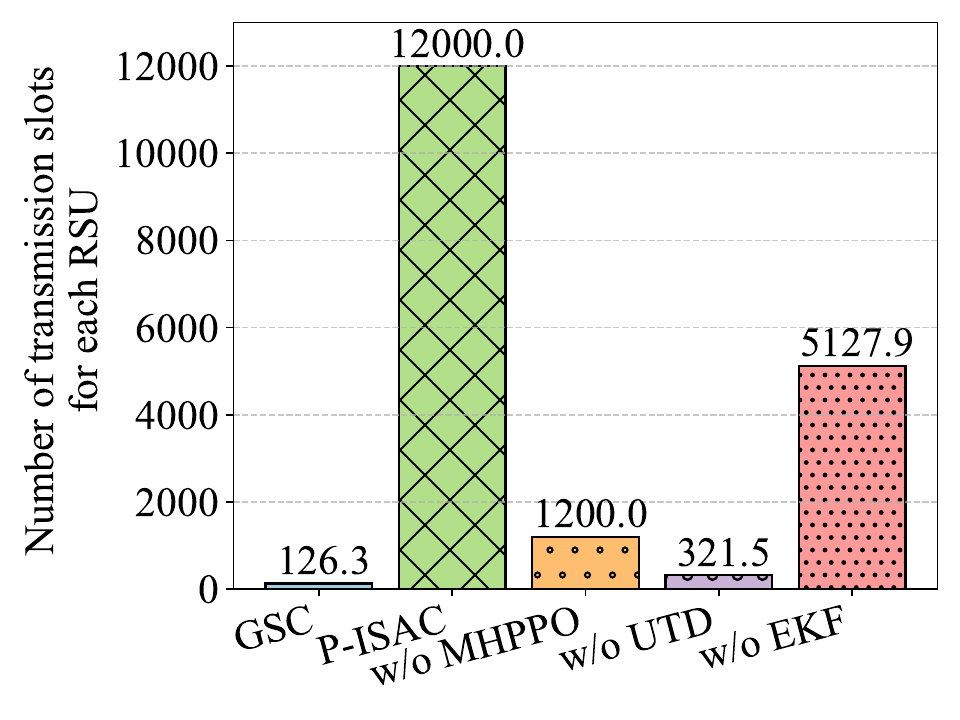}
        \label{trans_slot}
    } 
    \subfloat[][Communication success rate.]
    {\includegraphics[width=0.23\textwidth]{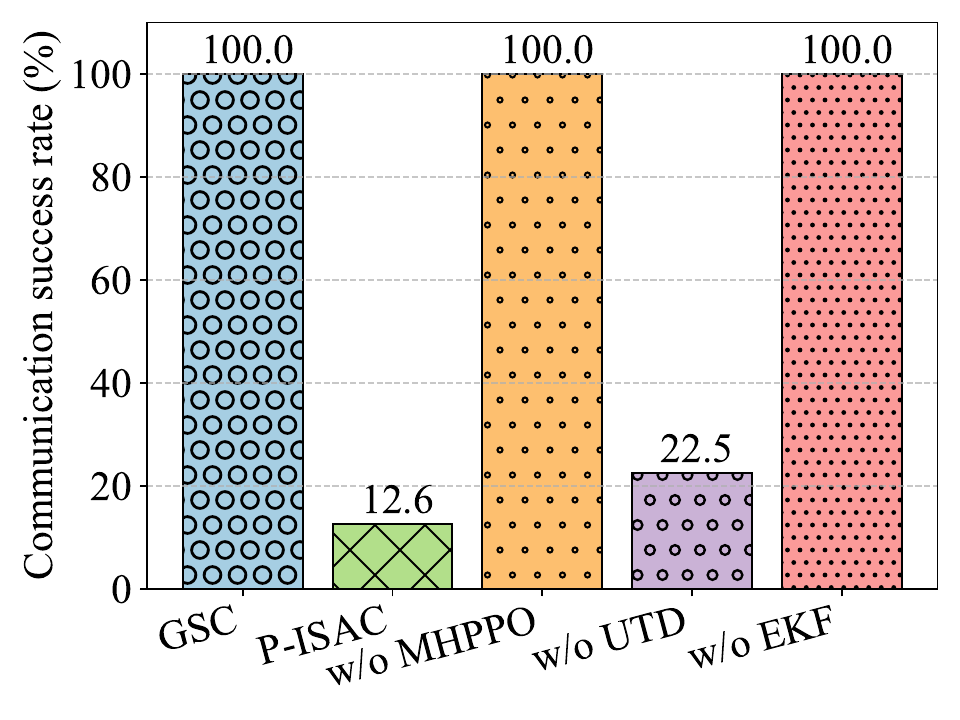}
        \label{com_rate}
    }
    \caption{Performances under different schemes.}
    \label{statistics}
    \vspace{-0.3 cm}
\end{figure}

To further examine objective-related performance, \figref{statistics} presents the total number of passed vehicles, the total number of transmitted signals for each vehicle, the total number of transmission slots for each RSU, and the communication success rate under different schemes, where only successful tasks are considered for fair comparison. As observed, our proposed GSC framework achieves 39.2 passed vehicles, requires only 12.9 transmitted signals for each vehicle and 126.3 transmission slots for each RSU, and maintains a 100\% communication success rate. This indicates that GSC can identify task-critical signals and transmit them reliably, thereby enabling efficient vehicle coordination with very low signaling overhead. In contrast, the adapted P-ISAC baseline scheme passes only 20.4 vehicles while requiring the largest number of transmissions, i.e., 4932.7 transmitted signals for each vehicle and 12000 transmission slots for each RSU. This is mainly because its communication success rate is only 12.6\% due to severe inter-RSU interference caused by simultaneous distributed transmissions. Although timely C\&C signals can be generated, they cannot be reliably decoded, leading to inefficient vehicle motion control.

The w/o MHPPO scheme achieves the largest number of passed vehicles, but this comes at the cost of the lowest task success rate. This is because periodic transmission ignores the semantic importance of sensing and C\&C signals as well as potential collision risks, leading to overly aggressive vehicle movements. Therefore, although more vehicles may pass the intersection in some successful episodes, the overall safety performance is significantly degraded. The w/o UTD scheme achieves the second-lowest signaling overhead, requiring only 50.6 transmitted signals per vehicle and 321.5 transmission slots per RSU to support 25.8 passed vehicles. However, without the UTD module, its communication success rate remains low, which limits both traffic efficiency and signaling efficiency. The w/o EKF scheme achieves only 22.7 passed vehicles while requiring 1797.8 transmitted signals per vehicle and 5127.9 transmission slots per RSU. Since vehicle states cannot be continuously predicted when sensing signals are not transmitted, the system must rely on frequent sensing updates to avoid stale state information, resulting in high signaling overhead.

\begin{figure}[t] 
    \centering
    \subfloat[][Total number of passed vehicles.]
    {\includegraphics[width=0.23\textwidth]{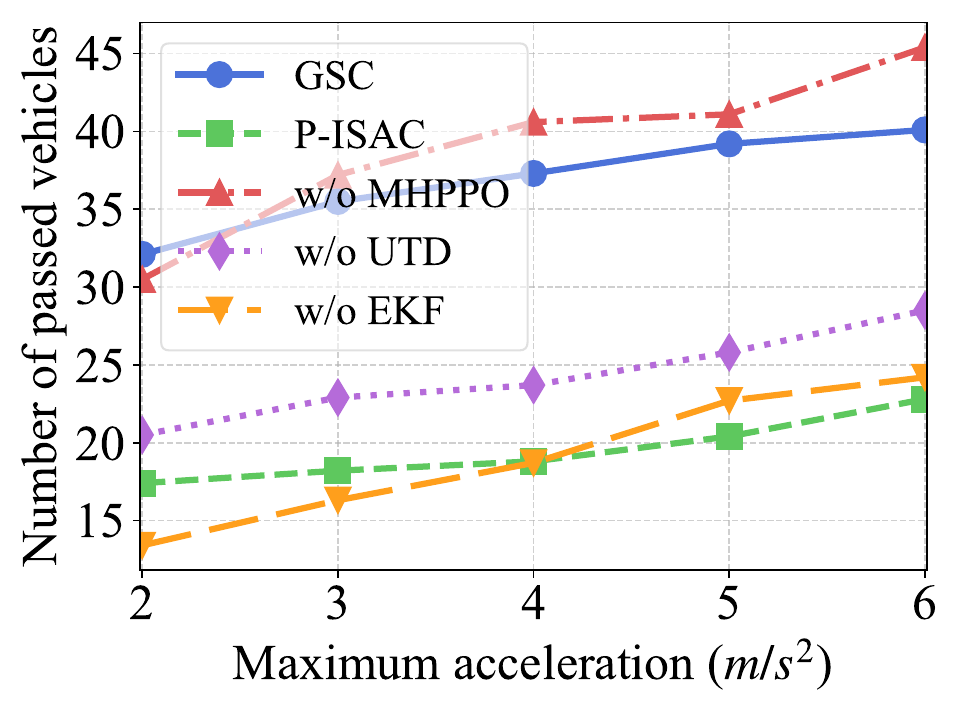}
        \label{vehicle_0}
    } 
    \subfloat[][Total number of transmitted signals for each vehicle.]
    {\includegraphics[width=0.23\textwidth]{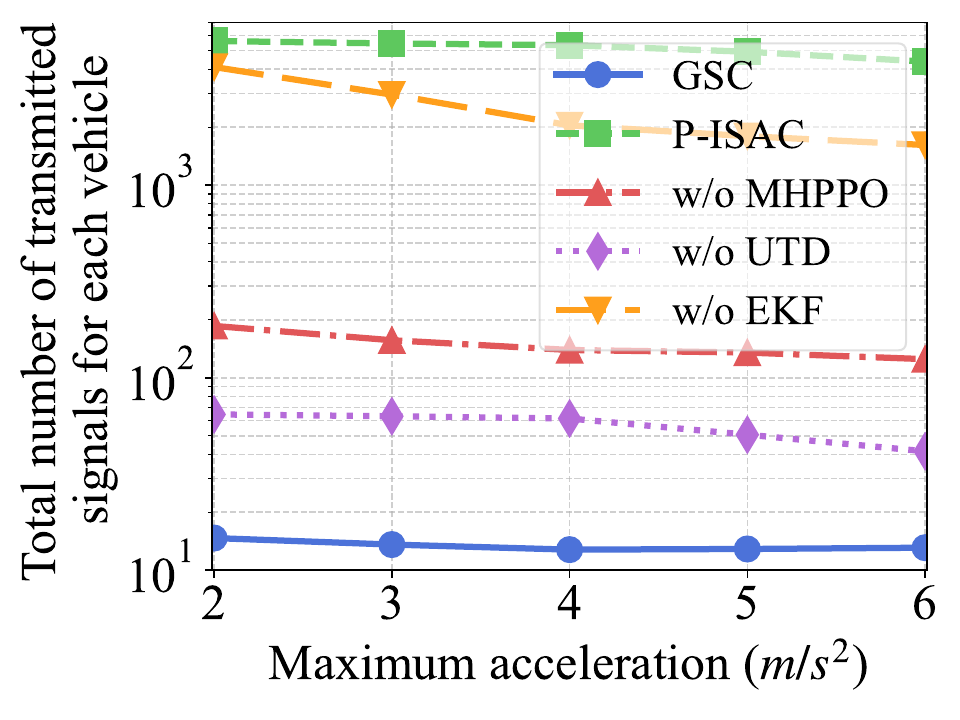}
        \label{signal_0}
    } 
    \caption{Performances under various maximum accelerations.}
    \label{acceleration_variation}
    \vspace{-0.3 cm}
\end{figure}   

\figref{acceleration_variation} plots the performance comparisons under different maximum accelerations. Note that only successful tasks are considered, and our proposed GSC framework maintains a 100\% task success rate across all simulations. Specifically, \figref{vehicle_0} presents the total number of passed vehicles. As the maximum acceleration increases, the number of passed vehicles generally increases for all schemes, since vehicles can adjust their speeds more flexibly and pass through the intersection more quickly. Although the w/o MHPPO scheme achieves the largest number of passed vehicles at high accelerations, this comes at the cost of a low task success rate.
\figref{signal_0} shows the total number of transmitted signals for each vehicle. As the maximum acceleration increases, the signaling overhead generally decreases. This is because vehicles with higher acceleration capabilities can complete intersection crossing within fewer time slots, thereby reducing the demand for sensing and C\&C signal transmissions. Our proposed GSC framework requires the fewest transmitted signals per vehicle under all acceleration settings. In particular, when the maximum acceleration exceeds 4 m/s$^2$, its signaling overhead remains nearly constant while the number of passed vehicles continues to increase. This demonstrates that GSC can exploit improved vehicle mobility to enhance traffic throughput without increasing signaling overhead, confirming its ability to efficiently schedule only task-critical sensing and C\&C transmissions.

\section{Conclusions}  \label{conclusion}
In this paper, we investigated distributed integrated sensing and communication (ISAC)-enabled vehicle coordination at intersections from a goal-oriented semantic communication (GSC) perspective. To improve signaling efficiency while ensuring safe vehicle coordination, we proposed a unified GSC framework that jointly integrates extended Kalman filter (EKF)-based vehicle state prediction, masked hybrid proximal policy optimization (MHPPO)-based value-of-information (VoI)-driven transmission scheduling, and uncertainty-aware transmission design (UTD). Specifically, the EKF enables continuous vehicle state prediction and distributed sensing fusion, MHPPO determines task-critical sensing and command and control (C\&C) signal transmissions, and UTD improves transmission reliability under vehicle state uncertainty and inter-cell interference.
Simulation results showed that our proposed GSC framework achieves 100\% collision-free vehicle coordination with significantly reduced signaling overhead compared with the adapted predictive ISAC baseline and ablation schemes. Our results further confirmed that the integration of EKF-based state prediction, MHPPO-based VoI-driven scheduling, and UTD-based transmission is essential for achieving safe and signaling-efficient distributed ISAC-enabled vehicle coordination.

\bibliographystyle{IEEEtran}
\bibliography{IEEEabrv,bib}

\end{document}